\documentclass[preprint,12pt]{elsarticle}

\usepackage{amssymb}
\usepackage{amsmath}
\usepackage{marvosym}
\usepackage{colortbl}
\usepackage{xcolor}
\usepackage{array}
\usepackage{diagbox}
\usepackage{algpseudocode} 
\usepackage[ruled]{algorithm2e}  
\usepackage{siunitx}
\usepackage{graphicx}
\usepackage{subfig}
\usepackage{hyperref}
\hypersetup{
	colorlinks=true,
	linkcolor=cyan,
	filecolor=blue,      
	urlcolor=red,
	citecolor=green,
}

\journal{  }
\begin{document}
\begin{frontmatter}

\title{Improved physics-informed neural network in mitigating gradient-related failures}

\author{Pancheng Niu \textsuperscript{a}} 
\author{Yongming Chen \textsuperscript{a}} 
\author{Jun Guo \textsuperscript{a, b, \Letter}}
\author{Yuqian Zhou \textsuperscript{a}} 
\author{Minfu Feng \textsuperscript{c}} 
\author{Yanchao Shi \textsuperscript{d}} 

\affiliation{organization={College of Applied Mathematics},
            addressline={Chengdu University of Information Technology}, 
            city={Chengdu},
            postcode={610225}, 
            state={Sichun},
            country={P. R. China}}

\affiliation{organization={Key Laboratory of Numerical Simulation of Sichuan Provincial Universities, School of Mathematics and Information Sciences},
            addressline={ Neijiang Normal Univeristy}, 
            city={Neijiang},
            postcode={641000}, 
            state={Sichun},
            country={P. R. China}}

\affiliation{organization={College of Mathematics},
            addressline={Sichuan University}, 
            city={Chengdu},
            postcode={610064}, 
            state={Sichun},
            country={P. R. China}}  
\affiliation{organization={School of Sciences},
            addressline={Southwest Petroleum University}, 
            city={Chengdu},
            postcode={610500}, 
            state={Sichun},
            country={P. R. China}}        
\begin{abstract}

Physics-informed neural networks (PINNs) integrate fundamental physical principles with advanced data-driven techniques, driving significant advancements in scientific computing. However, PINNs face persistent challenges with stiffness in gradient flow, which limits their predictive capabilities. This paper presents an improved PINN (I-PINN) to mitigate gradient-related failures. The core of I-PINN is to combine the respective strengths of neural networks with an improved architecture and adaptive weights containing upper bounds. The capability to enhance accuracy by at least one order of magnitude and accelerate convergence, without introducing  extra computational complexity relative to the baseline model, is achieved by I-PINN. Numerical experiments with a variety of benchmarks illustrate the improved accuracy and generalization of I-PINN. The supporting data and code are accessible at \url{https://github.com/PanChengN/I-PINN.git}, enabling broader research engagement.

\end{abstract}

\begin{graphicalabstract}
\centering
\includegraphics[scale=0.15]{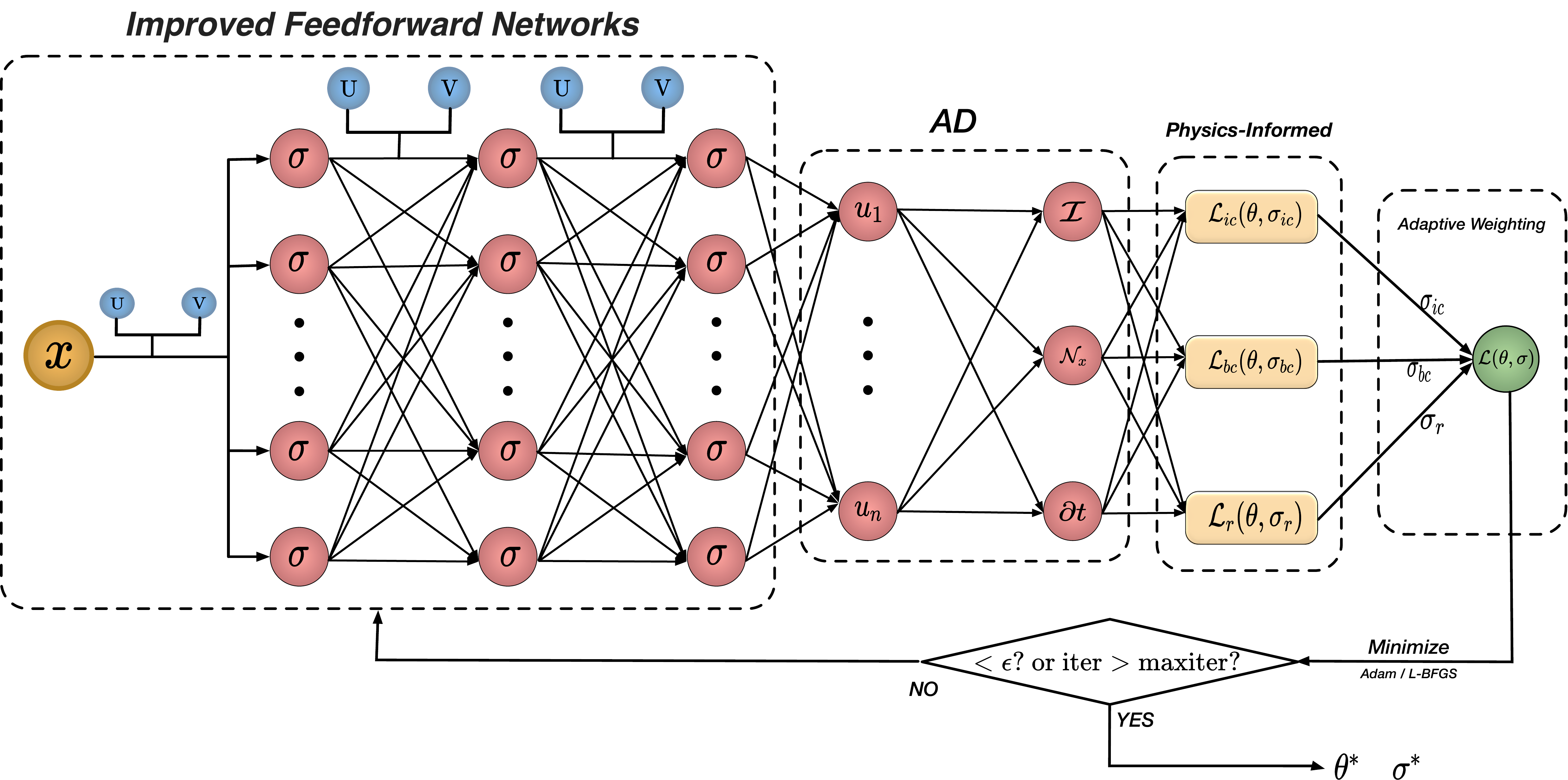}
\centering \includegraphics[width=0.9\textwidth]{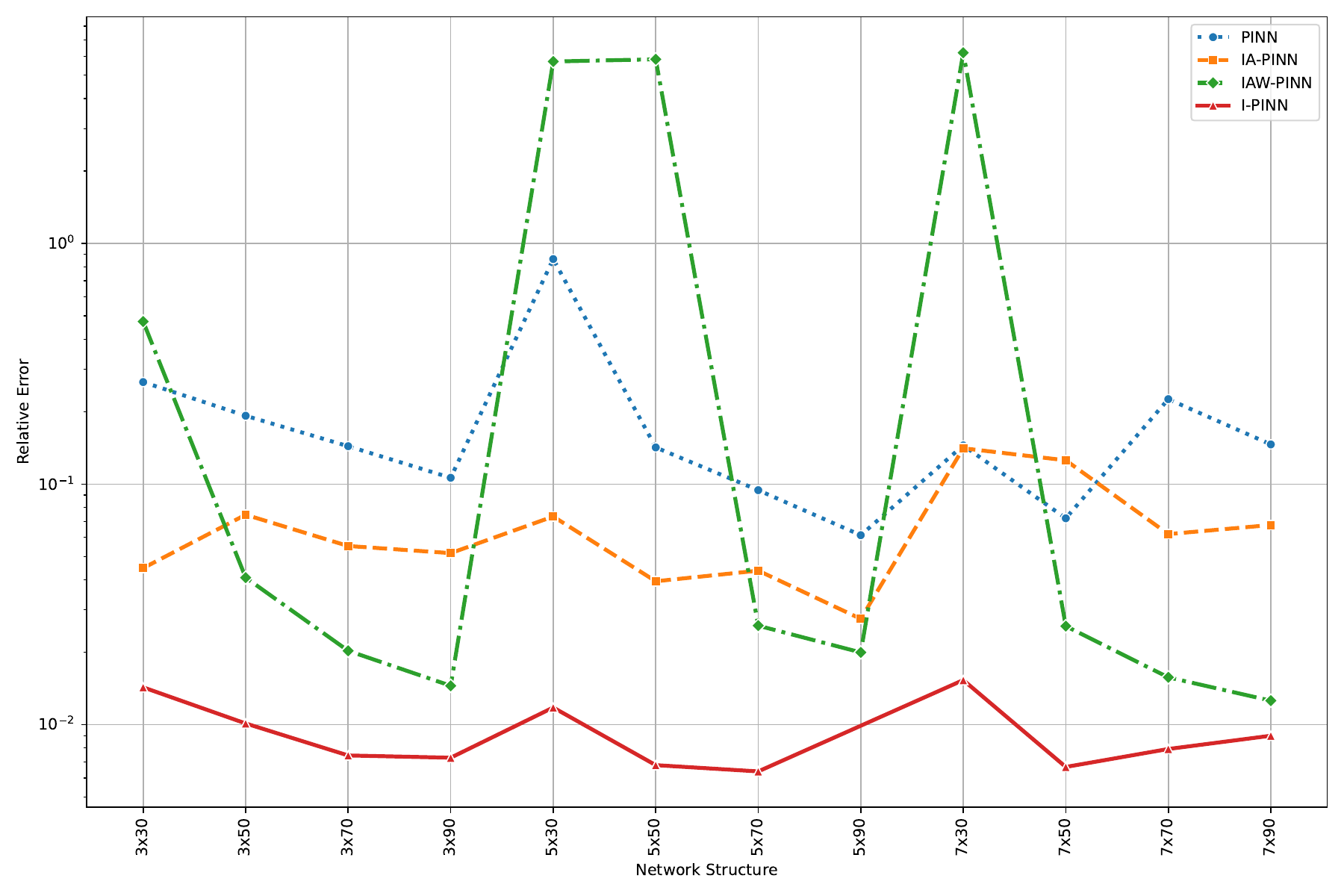}
\end{graphicalabstract}

\begin{highlights}

\item The uncertainty weighting method from \cite{arXiv211101394} has been enhanced with the Improved Adaptive Weighting PINN (IAW-PINN). This method sets an upper bound on adaptive weights, preventing excessive suppression of residual term weights and avoiding ill-conditioned optimization problems. It ensures convergence and adherence to constraints, improving both boundary condition and residual term accuracy, and thereby enhancing the overall stability and accuracy of I-PINN.
\item I-PINN effectively addresses the issues of substantial fluctuations in solution outcomes observed in IAW-PINN, significantly enhancing robustness across diverse network structures and improving the model’s generalization capabilities.
\item I-PINN significantly enhances solution accuracy, achieving improvements of at least an order of magnitude over PINN, IAW-PINN, and IA-PINN.
\end{highlights}

\begin{keyword}
Physics-informed neural networks\sep Gradient flow stiffness\sep Adaptive weighting\sep Scientific computing
\end{keyword}
\end{frontmatter}

\section{Introduction}
\label{sec1}

In the rapidly evolving fields of scientific computing and artificial intelligence, the integration of physical laws with data-driven models has given rise to a novel paradigm: Physics-Informed Neural Networks (PINNs) \cite{Karniadakis2021,MR3881695}. These advanced models extend beyond conventional deep learning frameworks by incorporating fundamental physical principles. This approach enables them to effectively address complex challenges across a wide range of disciplines, from fluid dynamics to quantum mechanics, with notable  potential \cite{MR4209661}. However, the effectiveness of PINNs relies on achieving a delicate balance among multiple loss terms (also referred to as stiffness in the gradient flow dynamics), each representing a distinct facet of the physical constraints imposed by partial differential equations (PDEs) \cite{MR4309866,MR4337814}.

In addressing this challenge, a plethora of adaptive modifications to  the vanilla  PINN framework has been devised. Given our current, albeit limited, understanding, these advancements can be systematically categorized into  four principal categories:
\begin{enumerate}
    \item The methods that focus on the refinement of the loss function aim to enhance the optimization process and convergence properties. For instance, the adaptive training weights ensure a balanced emphasis across training process \cite{arXiv211101394}; the min-max architectures balance error minimization with  robustness enhancement \cite{LIU2021112}; the mask functions impose selective constraints on input regions \cite{MR4746527}; the neural tangent kernel technique bolsters the learning capabilities of the network \cite{MR4337814}; and other innovative strategies that collectively contribute to the advancement of PINN \cite{MR4698526,MR4350501,MR4748739}.
    \item Dedicated to bolstering the network's representational capacity, these methods enable the network to adeptly capture intricate data patterns. This is achieved through enhancements such as sophisticated network architectures \cite{MR4309866} and adaptive activation functions that are tailored to the complexity of the underlying data \cite{MR4051868,MR4133779}.
    \item Adaptive configuration techniques dynamically adjust the network's parameters to better suit the problem at hand \cite{MR4426047,Hou2023,MR4209661,MR4499321}.
    \item Domain decomposition methods segment the problem space to address multi-scale challenges more effectively \cite{Hu_2023,MR4188528,MR4083367}.
\end{enumerate}

By employing robust optimization algorithms and meticulously tuning their parameters, these studies demonstrate that optimizing an objective function with loss terms of varying magnitudes can significantly enhance overall optimization performance. Nevertheless, achieving accurate predictions for engineering applications remains a formidable challenge \cite{MR4764284}, and no single existing method has yet proven fully adequate to address it.

The impetus for this paper is to develop an improved PINN in mitigating gradient-related failures, thereby enhancing predictive accuracy. In previous works, Wang et al. \cite{MR4309866} noted gradient pathologies in PINN , proposing an adaptive learning rate annealing to balance fitting and regularization. Also, they presented  an  improved fully connected neural architecture for  PINN (termed the improved architecture for PINN here, abbreviated as IA-PINN), designed to mitigate  stiffness issues. Huang et al. \cite{arXiv211101394} developed  an improved adaptive weights algorithm for PINN featuring a constrained lower bound on uncertainty weighting. This approach dynamically equalizes the impact of diverse loss components during training, applying a nonnegative floor to each term's uncertainty to ensure stable and convergent neural network optimization for PDEs. Kendall et al.'s work \cite{arXiv170507115} tends to encounter a scenario where one of the loss terms (typically the initial and boundary loss terms) consistently dominates during model training, this results in restrained growth of the weights associated with the residual terms and inadequate training of the unsupervised learning component. To this end, we  have enhanced the work of  Huang et al \cite{arXiv211101394} by transforming  the lower bound of uncertainty into the upper bound of the weight, thereby developing an improved adaptive weighting method referred to as IAW-PINN. This improvement allows the upper bound of the weight to be modified more intuitively and flexibly, thereby controlling the relative balance of weights among different loss terms and better promoting both the supervised and unsupervised learning of PINNs. However, we have also observed that the single method mentioned above is affected by a variety of factors, such as network structure and weighting, and therefore its accuracy needs to be further improved.

To address this problem, we have developed an improved PINN (I-PINN) by integrating the respective strengths of the IA-PINN and IAW-PINN. Through a series of representative numerical experiments, I-PINN has demonstrated the potential to reduce errors by at least an order of magnitude and improve the convergence rate, all while maintaining computational complexity comparable to the baseline model. Specifically, our principal contributions are as follows:
\begin{enumerate}
\item The uncertainty weighting method from \cite{arXiv211101394} has been enhanced with the IAW-PINN. IAW-PINN sets an upper bound on adaptive weights, preventing excessive suppression of residual term weights and avoiding ill-conditioned optimization problems \cite{arXiv210204626}. It ensures convergence and adherence to constraints, improving both boundary condition and residual term accuracy, and thereby enhancing the overall stability and accuracy of I-PINN.
\item I-PINN effectively addresses the issues of substantial fluctuations in solution outcomes observed in IAW-PINN, significantly enhancing robustness across diverse network structures and improving the model’s generalization capabilities.
\item I-PINN significantly enhances solution accuracy, achieving improvements of at least an order of magnitude over PINN, IAW-PINN, and IA-PINN.
\end{enumerate}

The paper is structured as follows: Section \ref{sub21} recalls the vanilla PINN. Sections \ref{sub22} and \ref{sub23} detail the neural architecture and uncertainty weighting, essential for addressing PINN challenges. In Section \ref{sub24}, we introduce an adaptive weighting model for PINN by integrating an enhanced architecture with uncertainty weighting, representing the core innovation of this research. Section \ref{sub3} provides a comprehensive evaluation of our models through benchmark cases, assessing their performance. The final section concludes with insights on limitations and future directions. To foster openness and reproducibility, all associated code and data are made accessible in the public domain at \url{https://github.com/PanChengN/I-PINN.git}.

\section{Methods}\label{sub2}
\subsection{ Vanilla PINN framework}\label{sub21}
In reference \cite{MR3881695}, the vanilla PINN is introduced as a method to solve partial differential equations using deep neural networks. The prediction function is derived by optimizing the loss function through various optimization algorithms. This loss function comprises a supervised term and a residual term. The supervised term measures the degree of approximation to the deterministic conditions, including boundary and initial conditions (if applicable). The residual term, obtained via automatic differentiation techniques, measures the degree of approximation to the governing equation. The PINN method is applicable to various types of partial differential equations, including fluid dynamics equations, heat conduction equations, electromagnetic equations, and others.

Here, we consider a generalized nonlinear partial differential equation:
\begin{equation}\label{1}
    \begin{cases}\mathcal{D}(u(x,t))=f(x,t),&x\in\Omega\subset\mathbb{R}^d,\\\mathcal{B}(u(x,t))=g(x,t),&x\in\partial\Omega,\\\mathcal{I}(u(x,0)=h(x), &x\in\Omega,
    \end{cases}
\end{equation}
where $\mathcal{D}$ denotes the differential operator,  $\mathcal{I}$ denotes the initial condition operator, and $\mathcal{B}$ is the boundary condition operator, such as the Dirichlet, Neumann, Robin, or Periodic condition. $u(x, t)$ is the solution function on the $d$-dimensional domain $\Omega$, $f(x,t)$ is a given source or forcing function that introduces external influences into the system.


The loss function in PINN is designed to contain the deviation of the neural network solution from the initial/boundary conditions, the partial differential equation itself. Weighted multipliers $ \lambda_{ic} $, $ \lambda_{bc}$, and $ \lambda_r $ are used to balance the effects of these components in the loss function:
\begin{equation}
    \mathcal{L}=\lambda_{ic}\mathcal{L}_{ic}+\lambda_{bc}\mathcal{L}_{bc}+\lambda_{r}\mathcal{L}_{r}
\end{equation}
where $\mathcal{L}_{ic}$,  $\mathcal{L}_{bc}$ and $ \mathcal{L}_{r}$ denote the loss terms for the initial conditions, boundary conditions, and the residuals associated with the PDE, respectively. These terms are given by the following:
\begin{equation}
    \mathcal{L}_{ic}=\frac{1}{N_{\mathrm{ic}}}\sum_{i=1}^{N_{\mathrm{ic}}}|\mathcal{I}(\hat{u}(x_{i,\mathrm{ic}}))-h(x_{i,\mathrm{ic}})|^{2}
\end{equation}
\begin{equation}
     \mathcal{L}_{bc}=\frac{1}{N_{\mathrm{bc}}}\sum_{i=1}^{N_{\mathrm{bc}}}|\mathcal{B}(\hat{u}(x_{i,\mathrm{bc}}, t_{i,\mathrm{bc}}))-g(x_{i,\mathrm{bc}}, t_{i,\mathrm{bc}})|^{2}
\end{equation}
\begin{equation}
    \mathcal{L}_{r}=\frac{1}{N_{\mathrm{r}}}\sum_{i=1}^{N_{\mathrm{r}}}|\mathcal{D}(\hat{u}(x_{i,\mathrm{r}}, t_{i,\mathrm{r}}))-f(x_{i,\mathrm{r}}, t_{i,\mathrm{r}})|^{2}
\end{equation}
where $\hat{u}(x,t;\theta)$ is the neural network prediction, $x_{i,j}$ is the spatio-temporal coordinates,  $N_{ic}$, $N_{bc}$, and $N_r$ are the collocation of initial, boundary, and residual points. Initial point data is used to enforce initial conditions on the neural network, while boundary point data is used to ensure that the neural network satisfies the boundary conditions. Internal collocation points are randomly selected coordinate points from the domain and are used to compute the residual loss, which forces the neural network $ \hat{u}(x,t;\theta)$ to satisfy the governing equations. After preparing the training data, we can optimize the neural network parameters using gradient descent methods commonly used in deep learning, such as Adam, SGD, or L-BFGS. By minimizing the loss function and making $\mathcal{L}$ as close to zero as possible, we can consider the neural network $ \hat{u}(x,t;\theta) $ as an approximate solution function $ u(x,t)$ of (\ref{1}) when the loss reaches a minimum.

\subsection{Improved Multilayer perceptron}\label{sub22}
In recent studies, in order to alleviate the gradient vanishing/explosion problem of the original multilayer perceptron when processing complex problems \cite{MR4425212}, it has been shown that PINNs are not well-suited for handling such issues. A recently improved fully connected feedforward neural network \cite{MR4309866}, inspired by the prominent attention mechanism in Transformer \cite{MR4550083}, effectively addresses the limitations encountered by the vanilla PINN. This enhanced network improves the performance of PINNs by embedding the input variable \( x \) into the hidden state of the network. Initially, these inputs are encoded in the feature space using two different encoders, \( U \) and \( V \), as given by the following formula:
\begin{equation}
    U=\sigma(x^0\theta^U+b^U),\quad V=\sigma(x^0\theta^V+b^V)
\end{equation}
The encoder is then assimilated into each hidden layer of the conventional MLP by point-by-point multiplication. Thus, each forward propagation becomes:

\begin{align}
&\alpha ^{l}( x) = \alpha ^{l- 1}( x) \theta ^{l}+ b^{l}, \mathrm{~for~} l\in \{ 1, 2, \ldots , N\}\\
&\alpha^l(x)=\sigma(\alpha^l(x))\\
&\alpha^{l}(x)=(1-\alpha^{l}(x))\odot U+\alpha^{l}(x)\odot V
\end{align}
where $x$ is the input, $\alpha^l$ and $\theta^l$ are the neurons and weights of the $l$th layer, $\sigma$ is the activation function, and $\odot$ is the element-by-element product.

\subsection{Adaptive loss weight functions based on weight upper bounds.}\label{sub23}

In the training of vanilla PINN, it is well-recognized that while it effectively solves many simple problems, it often struggles with more complex problems. A significant contributing factor to this limitation is the fixed weighting of the loss function. In vanilla PINN, the weight $\lambda_i$ assigned to the loss function is typically set to 1 and remains unchanged throughout training.

Hence, selecting appropriate weights $\lambda_{ic}$, $\lambda_{bc}$, and $\lambda_r$ to balance different loss components can significantly accelerate the convergence of the PINNs model and enhance its accuracy \cite{MR4513793,Xiang_Peng_Liu_Yao_2022}.

In this paper, we adopt the multi-task learning approach as outlined in \cite{Cipolla_Gal_Kendall_2018}. To modify the weights of the loss function, we utilize the homoskedastic uncertainty associated with each task. Specifically, this method adjusts the adaptive weights through maximum likelihood estimation, allowing for adaptive weighting of different components of the loss. The specific procedure for weighting the adaptive loss function is detailed as follows.

We establish a Gaussian probabilistic model with output $u$. The Gaussian likelihood is defined as Gaussian with mean given by the approximation of PINNs $\hat{u}(x, t; \theta)$ and the uncertainty parameter $\sigma_m$:
\begin{equation}
    p(u|\hat{u}(x,t,\theta))=\mathcal{N}(u|\hat{u}(x,t,\theta),\sigma_m^2)
\end{equation}
where, $\sigma_m$ is a noise parameter representing the uncertainty of the $m$ th task. Assuming a total of $M$ tasks, we maximise the joint probability distribution over all tasks.
\begin{equation}
    p(u_1,u_2,...,u_M|\hat{u}(x,t,\theta))=\prod_{m=1}^Mp(u_m|\hat{u}(x,t,\theta))
\end{equation}
Then, we take the negative logarithm of the likelihood function so that we can obtain the negative logarithmic likelihood as the desired loss function.

\begin{align}
&-\log p(u_1,u_2,...,u_M|\hat{u}(x,t,\theta))\\
&=-\sum_{m=1}^M\log p(u_m|\hat{u}(x,t,\theta))\\
&=-\sum_{m=1}^M\log\mathcal{N}(u_m|\hat{u}(x,t,\theta),\sigma_m^2)\\
&\propto\quad\sum_{m=1}^M\left(\frac1{2\sigma_m^2}\mathcal{L}_m+\frac12\log\sigma_m^2\right)\label{15}
\end{align}

The first term in (\ref{15}) represents the weighted loss function for each task and the second term is a regularisation term to penalise tasks with high uncertainty. Minimising (\ref{15}) is equivalent to maximising the joint probability distribution of all the losses, which enables the adaptive assignment of appropriate weighting coefficients to each portion of the losses so that its various components are balanced. The parameter $\sigma_m$ represents the uncertainty of each task, if an item is easy to learn and has low uncertainty, it will be assigned a higher weight, on the contrary, tasks with high uncertainty will have a lower weight. At the same time, the regularisation term ensures that $\sigma_m$ does not become too large, preventing the simulation from ignoring difficult tasks.

In summary, the uncertainty-based adaptive loss function is
\begin{equation}\label{16}
    \begin{aligned}\mathcal{L}(\theta,\sigma)&=\frac1{2\sigma_{ic}^2}\mathcal{L}_{ic}(\theta)+\frac1{2\sigma_{bc}^2}\mathcal{L}_{bc}(\theta)+\frac1{2\sigma_r^2}\mathcal{L}_r(\theta)\\&+\frac12\log\sigma_i^2+\frac12\log\sigma_b^2+\frac12\log\sigma_r^2\end{aligned}
\end{equation}
Since (\ref{16}) is prone to a scenario where one of the losses (typically the initial and boundary loss terms) consistently dominates during model training, resulting in restrained growth of the weights associated with the residual terms and inadequate training of the unsupervised learning component, we propose an adaptive loss weight function incorporating upper bounds, inspired by \cite{arXiv211101394}. Specifically, we introduce a weight upper bound $\gamma$ to (\ref{16}), transforming it into the following equation.

\begin{align}
    \mathcal{L}(\theta, \sigma) &= \frac{1}{2(\sigma_{ic}^2 + \gamma^{-1})}\mathcal{L}_{ic}(\theta) + \frac{1}{2(\sigma_{bc}^2 + \gamma^{-1})}\mathcal{L}_{bc}(\theta) + \frac{1}{2(\sigma_r^2 + \gamma^{-1})}\mathcal{L}_r(\theta) \nonumber \\
    &\quad + \frac{1}{2}\log(\sigma_{ic}^2 + \gamma^{-1}) + \frac{1}{2}\log(\sigma_{bc}^2 + \gamma^{-1}) + \frac{1}{2}\log(\sigma_r^2 + \gamma^{-1}) \label{17}
\end{align}

\begin{align}
    &\propto \frac{1}{\sigma_{ic}^2 + \gamma^{-1}}\mathcal{L}_{ic}(\theta) + \frac{1}{\sigma_{bc}^2 + \gamma^{-1}}\mathcal{L}_{bc}(\theta) + \frac{1}{\sigma_r^2 + \gamma^{-1}}\mathcal{L}_r(\theta) \nonumber \\
    &\quad + \log(\sigma_{ic}^2 + \gamma^{-1}) + \log(\sigma_{bc}^2 + \gamma^{-1}) + \log(\sigma_r^2 + \gamma^{-1}) \label{18}
\end{align}
Incorporating the upper bound $\gamma$ on the weights serves two purposes. Firstly, it prevents the weight term $\frac{1}{2\sigma_m^{2}}$ from diminishing to zero. In the conventional adaptive weighting method \cite{ Hou2023, Xiang_Peng_Liu_Yao_2022}(abbreviated as AW), $\frac{1}{2\sigma_m^{2}}$ undergoes a continuous increase as the uncertainty factor diminishes towards zero. Secondly, as the uncertainty factor approaches zero, the upper bound on the weights constrains their magnitude, ensuring that they do not become excessively large. This approach effectively prevents any single loss term, such as the initial/boundary loss, from dominating. Consequently, both the initial/boundary loss term and the residual term can be optimally trained, thereby enhancing overall model performance.

The final loss function will be given by (\ref{18}), where $\sigma = \{\sigma_{ic}, \sigma_{bc}, \sigma_r\}$ are trainable parameters implicitly integrated into the model. These parameters can be optimized using gradient-based optimizers such as Adam or L-BFGS. In contrast to the standard PINNs model, $\lambda_m$ is defined as $\frac{1}{\sigma_m^2 + \gamma^{-1}}$.

In order to show that the IAW-PINN is more effective than the AW-PINN method,, we will use the example of the Helmholtz equation and conduct a comparative test between (\ref{16}) and (\ref{18}). For details on the setup of the Helmholtz equation, please refer to subsection \ref{sub31}.

\begin{figure}[htbp]
    \centering
    \includegraphics[width=0.8\linewidth]{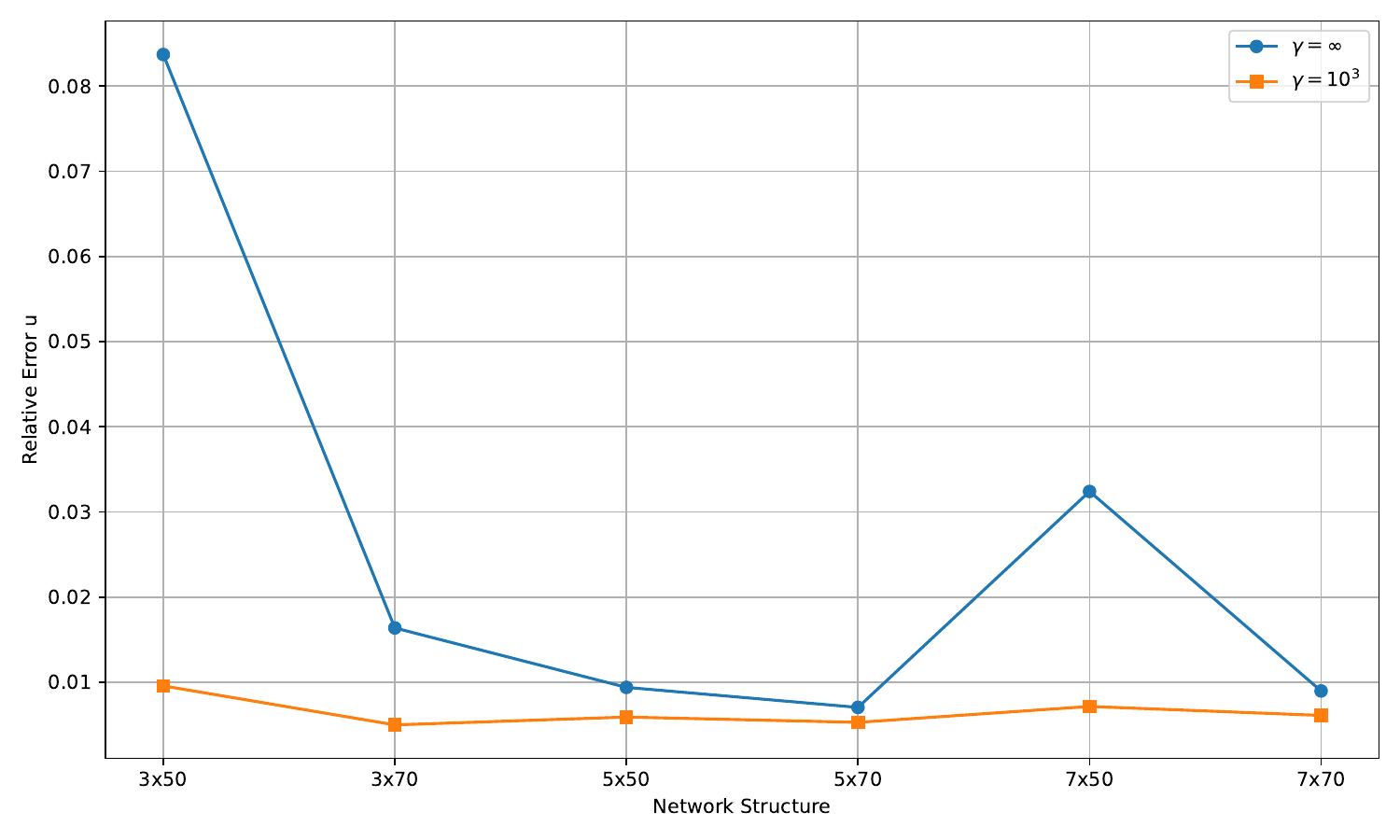}
    \caption{Comparison of the relative errors in solving \( u \) using different network structures between the AW-PINN and the IAW-PINN with upper bound \(\gamma = 10^3\)}
    \label{fig:relative_errors_u}
\end{figure}

\begin{figure}[htbp]
    \centering
    \includegraphics[width=0.8\linewidth]{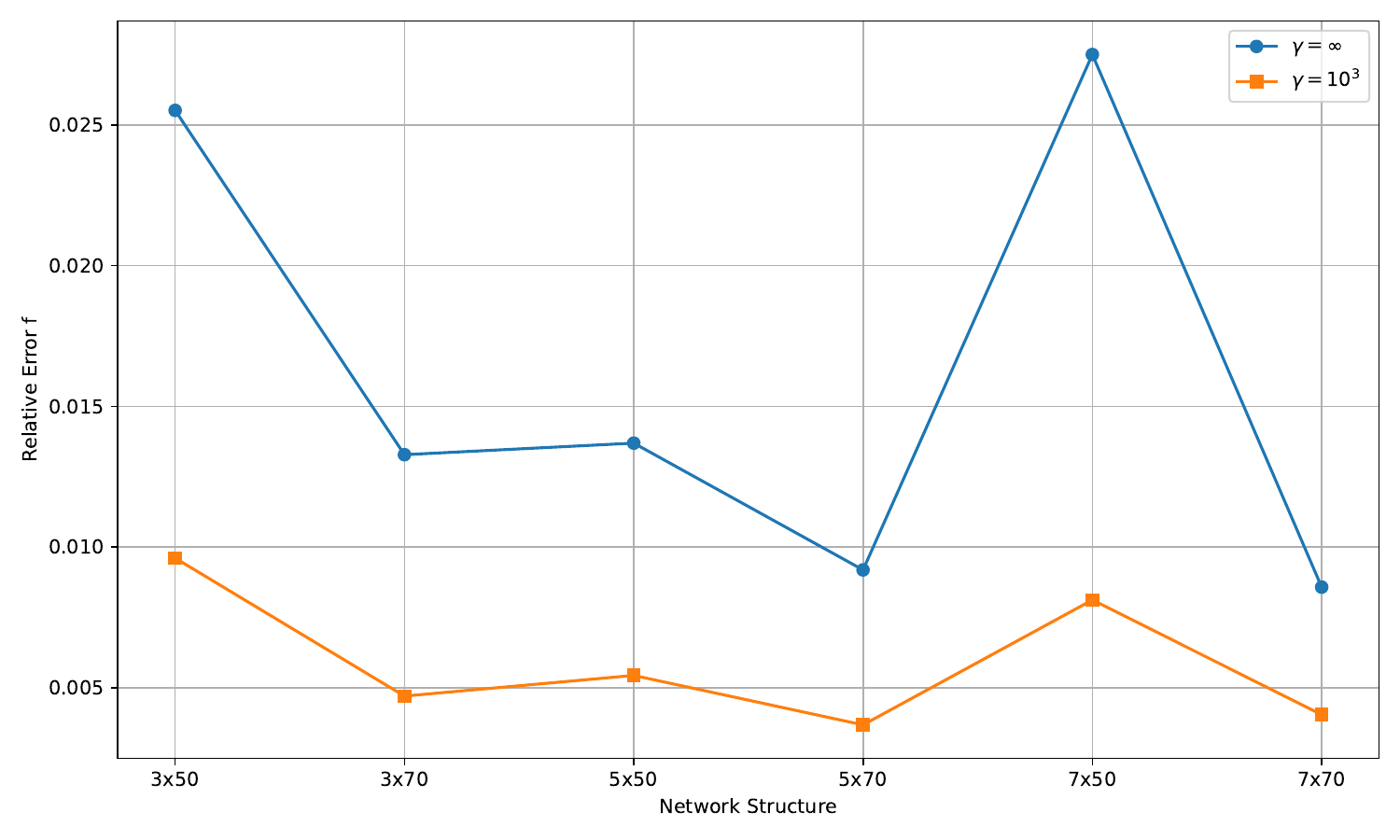}
    \caption{Comparison of the relative errors in solving \( f \) using different network structures between the AW-PINN and the IAW-PINN with upper bound \(\gamma = 10^3\)}
    \label{fig:relative_errors_f}
\end{figure}

\begin{figure}[htbp]
    \centering
    \includegraphics[width=0.8\linewidth]{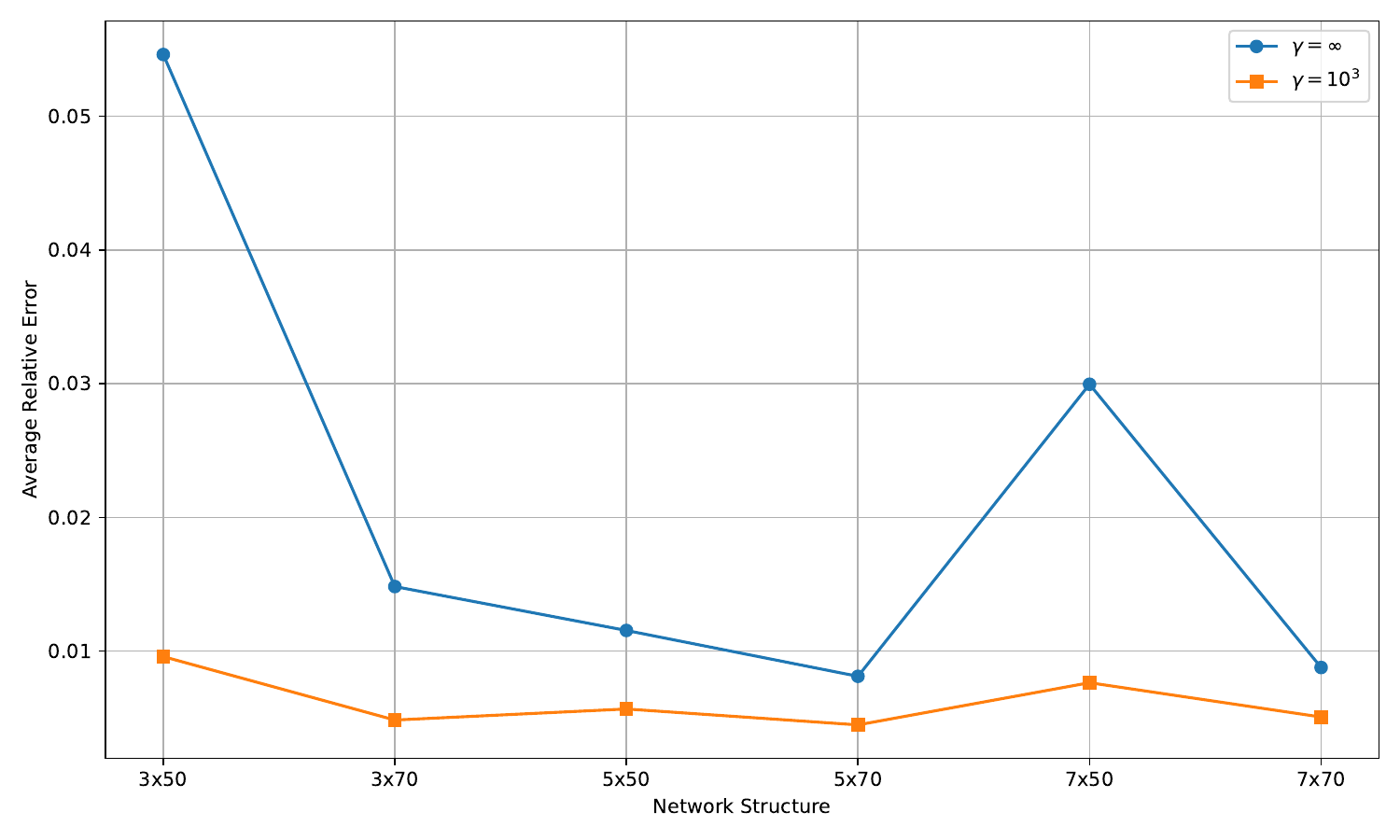}
    \caption{Comparison of the relative errors in solving \( u \) and \( f \) using different network structures between the AW-PINN and the IAW-PINN with upper bound \(\gamma = 10^3\)}
    \label{fig:average_relative_errors}
\end{figure}

It is evident from Figure \ref{fig:relative_errors_u} and Figure \ref{fig:relative_errors_f} that the weight-based upper bound IAW-PINN outperforms the vanilla adaptive loss weighting method \cite{Hou2023, Xiang_Peng_Liu_Yao_2022} in terms of both the solution \( u \) and the residual term \( f \) (unsupervised learning part). In summary, this approach effectively reduces the overall error of the model, allowing it to effectively address both supervised and unsupervised learning aspects. This is further illustrated in Figure  \ref{fig:average_relative_errors}, demonstrating superior model solving performance.

\begin{figure}[htbp]
    \centering
    \includegraphics[width=0.7\linewidth]{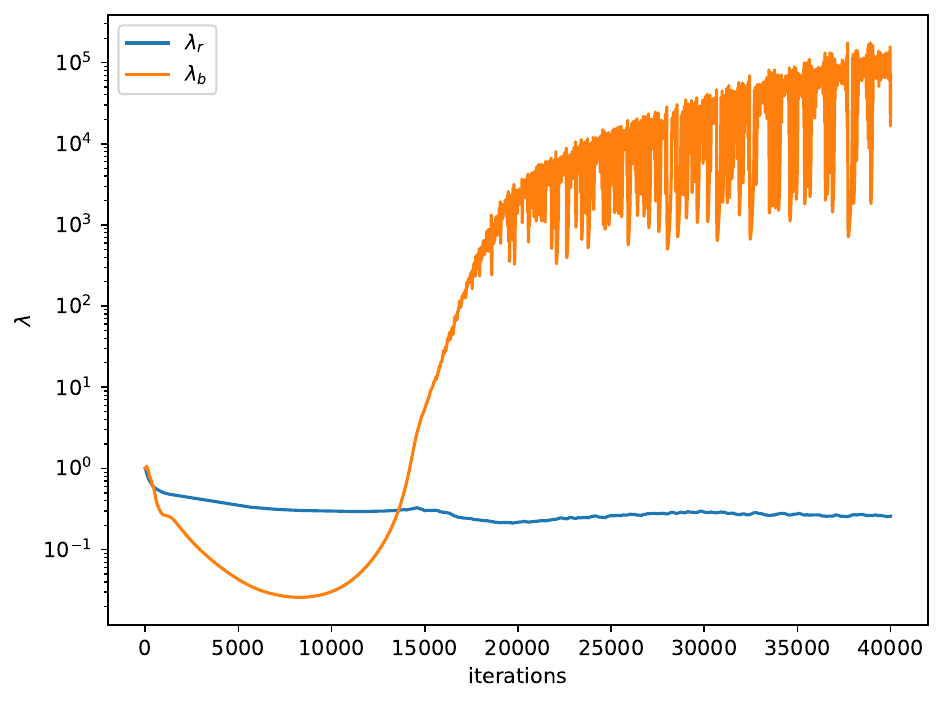}
    \caption{The figure shows the progression of loss function weights over 40,000 iterations of network optimization using Adam's optimizer, within a hidden layer network structure of \(7 \times 50\), employing the AW-PINN.}
    \label{fig:no IAW-PINN lam_log}
\end{figure}

\begin{figure}[htbp]
    \centering
    \includegraphics[width=0.7\linewidth]{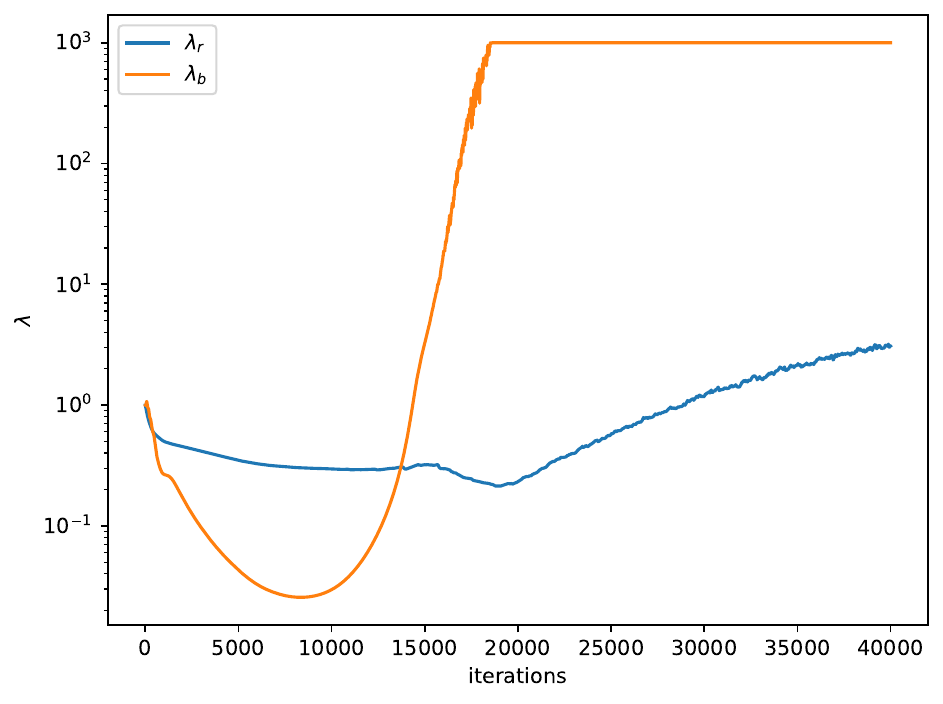}
    \caption{Variation of loss function weights obtained by optimizing the Network 40,000 times with Adam's optimizer under a hidden layer network structure \(7 \times 50\) using IAW-PINN with upper bound \(\gamma = 10^3\)}
    \label{fig:IAW-PINN lam_log}
\end{figure}

We observe from Figure \ref{fig:no IAW-PINN lam_log} and Figure \ref{fig:IAW-PINN lam_log} that the IAW-PINN effectively manages the weights of the boundary term (supervised part) and the residual term (unsupervised part). In Figure \ref{fig:no IAW-PINN lam_log}, the rapid increase in the weight of the boundary loss term not only suppresses the growth of the residual loss term's weight, leading to inadequate training of the residual term, but also introduces a significant disparity in the magnitudes of these two weights. This disparity can create optimization challenges and adversely impact the final solution quality. In contrast, Figure \ref{fig:IAW-PINN lam_log} illustrates that the IAW-PINN successfully mitigates these issues.

\subsection{Improved PINN framework}\label{sub24}
\begin{figure}[htbp]
    \centering
    \includegraphics[scale=0.15]{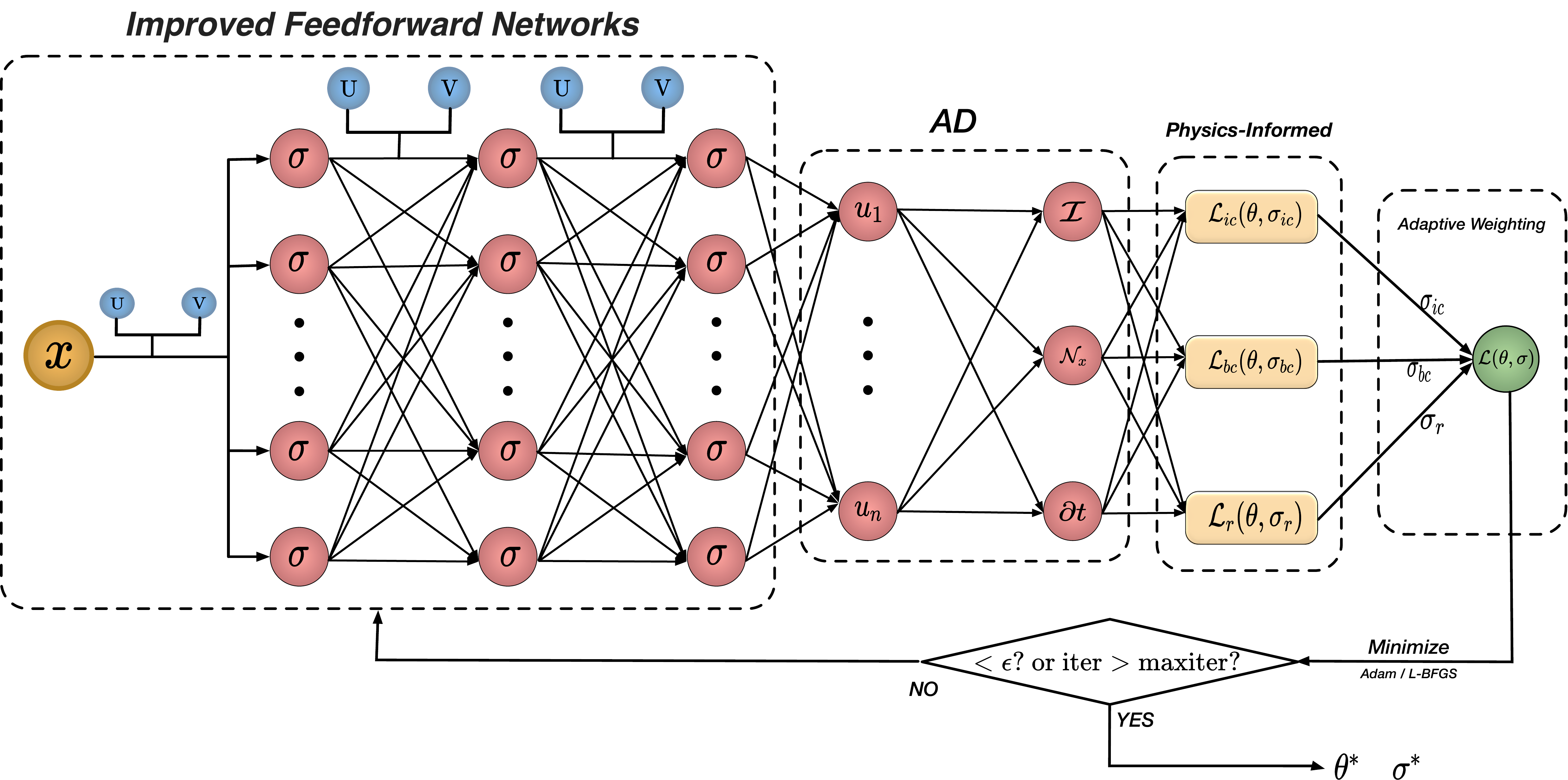}
    \caption{I-PINN framework}
    \label{fig:I-PINN framework}
\end{figure}

Network frameworks play a crucial role in the effective implementation of PINN methods in scientific computing. Addressing challenges such as vanishing/exploding gradients during PINN backpropagation in complex problems and mitigating the issue of imbalanced training due to scale differences in multi-objective losses is essential. To tackle these challenges, we propose a novel framework by integrating IA-PINN and IAW-PINN, which yields superior results compared to either method alone. The enhanced PINN framework, illustrated in Figure \ref{fig:I-PINN framework}, integrates an improved network structure with adaptive loss weighting methods. This approach is referred to as the improved Physical Information Neural Network (I-PINN). Algorithm \ref{alg:I-PINN algorithm} provides the pseudocode for the I-PINN algorithm.

\begin{algorithm}[H]
        \caption{I-PINN algorithm}
        \label{alg:I-PINN algorithm}
        \KwIn{$\sigma_{ic}, \sigma_{bc}, \sigma_{r}, \gamma, N_{ic}, N_{bc}, N_{r}$, epoch}
        \KwOut{$\theta^*, \sigma_i^*$}
         
        Initialize the adaptive weight parameters $\sigma_{ic}, \sigma_{bc}, \sigma_{r}$, and learning rate $\eta$; \\
        \For{i = 0, 1, ..., epoch }{
            $\theta_{i+1} = Adam \quad\mathcal{L}(\theta_i, \sigma_i)$;\\
            $\sigma_{i+1} = Adam \quad\mathcal{L}(\theta_i, \sigma_i)$;
            }
        Train the model with L-BFGS optimizer until convergence;\\
        \If{ error $<$ $\epsilon$}{
        \quad break;
        }
        \Return {$\theta^*$, $\sigma^*$}
\end{algorithm}

\section{Numerical examples}\label{sub3}
In this study, we systematically compare the PINN, IA-PINN, IAW-PINN and I-PINN in this paper using several numerical experiments. In order to verify the effectiveness of this PINN learning framework proposed in this paper.

\subsection{Helmholtz equation}\label{sub31}
The partial differential equation in two dimensions is expressed as follows:

\begin{equation}
    \dfrac{\partial^2u}{\partial x^2}+\dfrac{\partial^2u}{\partial y^2}+k^2u-q(x,y)=0   
\end{equation}

\begin{equation}
    \begin{aligned}
        q(x,y)&=-\:(a_{1}\pi)^{2}sin(a_{1}\pi x)sin(a_{2}\pi y)-(a_{2}\pi)^{2}sin(a_{1}\pi x)sin(a_{2}\pi y)\\ &+ k sin(a_{1}\pi x)sin(a_{2}\pi y)
    \end{aligned}
\end{equation}
where $q(x, y) $ is the forcing term that derives the analytic solution $u(x, y) = sin(a_1\pi x)sin(a_2\pi y)$. The boundary conditions are defined as follows:
\begin{eqnarray}
    u(-1, y) = u(1, y) = u(x, -1) = u(x, -1) = 0
\end{eqnarray}
Here, we adopt the configuration outlined by Wang et al. \cite{MR4309866}, with parameters set as \( a_1 = 1 \), \( a_2 = 4 \), and \( k = 1 \). We employ the Improved PINN framework (I-PINN) to systematically compare against Vanilla PINN,  IAW-PINN, and IA-PINN, aiming to demonstrate the effectiveness of our improved approach.

\begin{figure}[htbp]
    \centering
    \subfloat[PINN]{
        \includegraphics[width=\textwidth]{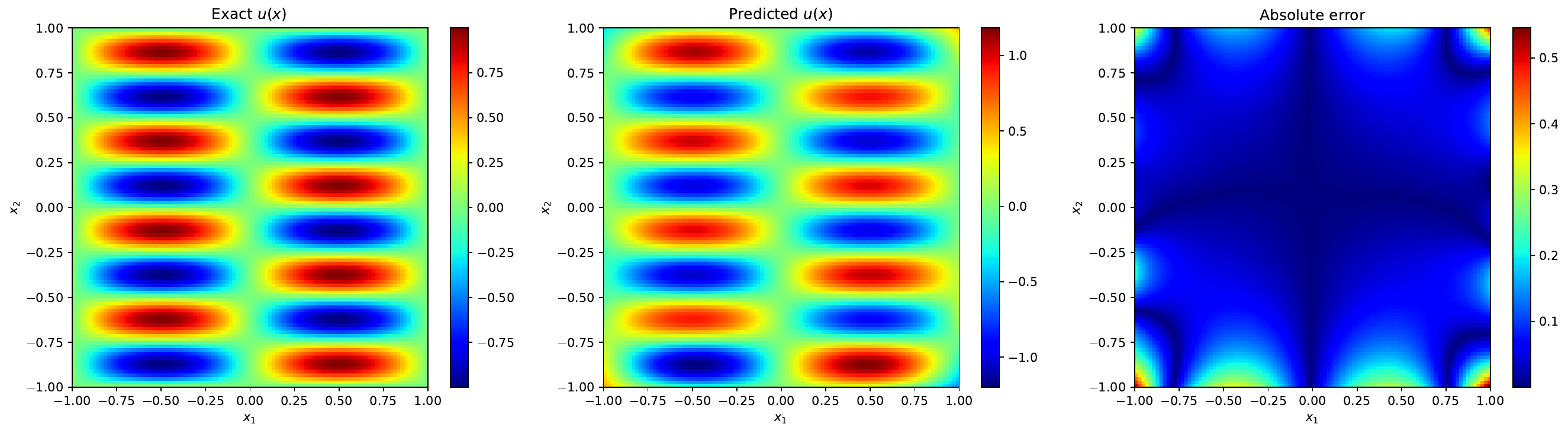}
    }\\
    \subfloat[IA-PINN]{
        \includegraphics[width=\textwidth]{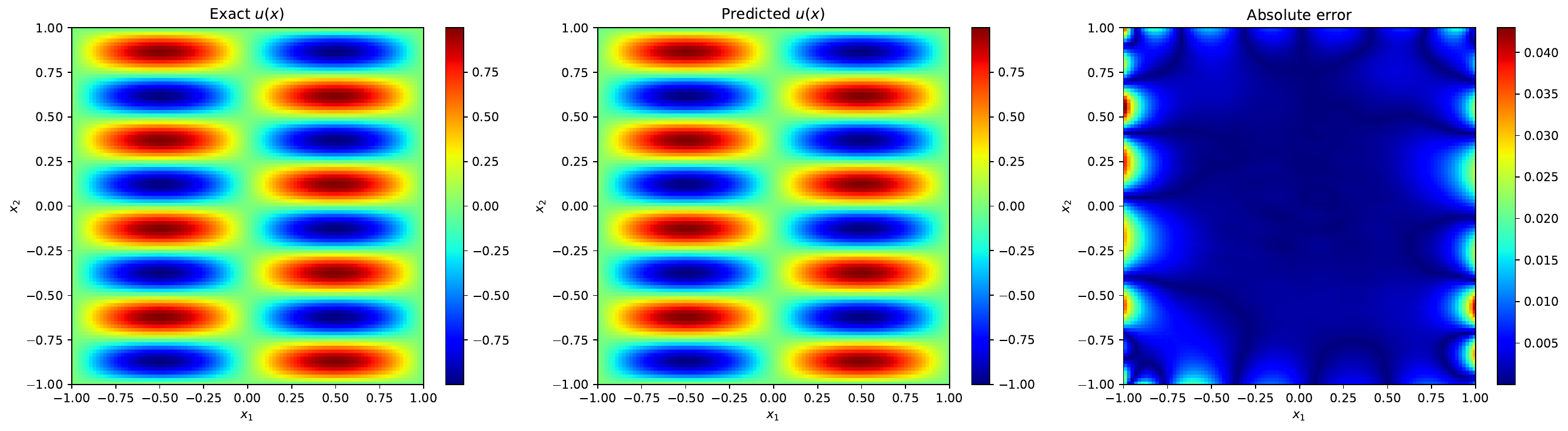}
    }\\
    \subfloat[IAW-PINN]{
        \includegraphics[width=\textwidth]{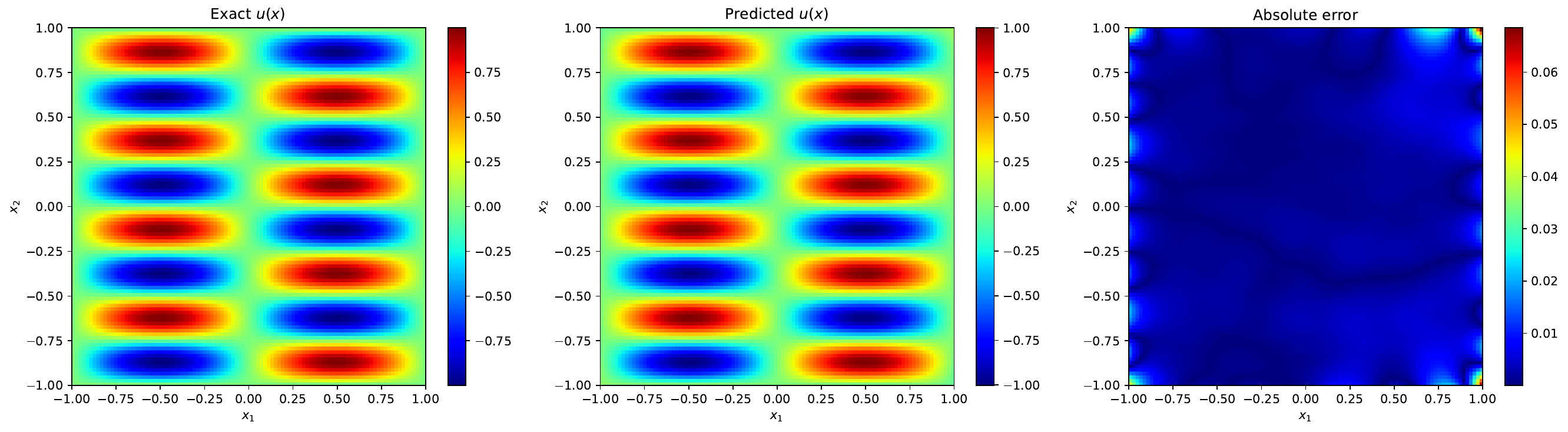}
    }\\
    \subfloat[I-PINN]{
        \includegraphics[width=\textwidth]{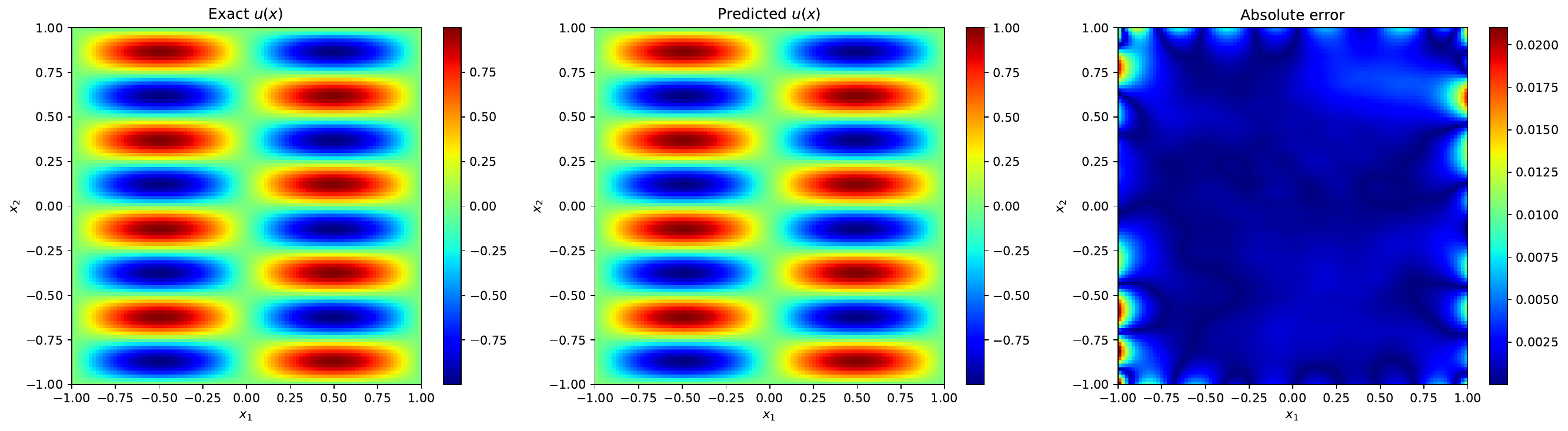}
    }
    \caption{From top to bottom, the figures show thermograms presenting the exact solution, predicted solution, and absolute error for PINN, IA-PINN, IAW-PINN, and I-PINN. And the corresponding relative errors are 2.68e-01, 2.22e-02, 1.21e-02, and 5.26e-03.}
    \label{fig:duibi}
\end{figure}

In this experiment, adaptive weights are initialized to \( 1 \). Training data comprises \( 512 \) boundary points and \( 128 \) interior points, processed over \( 40,000 \) iterations using small batches. Boundary points consist of \( 128 \) randomly sampled points from each of the four boundaries, while interior points are also randomly selected. The network architecture includes \( 7 \times 50 \) hidden layers. Optimization of PINN, IA-PINN, IAW-PINN, and I-PINN is performed using Adam optimizer with a maximum weight constraint of \( 100 \). Figure  \ref{fig:duibi} presents comparative results across different models, depicting exact solutions, predicted solutions, and absolute errors.

In this experiment, the relative errors obtained are as follows: PINN 2.68e-1, IA-PINN 2.22e-2, IAW-PINN 1.21e-2, and I-PINN 5.26e-3. Furthermore, comparative experiments were performed utilizing various network architectures, as detailed in Table \ref{tab:relative_errors}.

\begin{table}[h]
\centering
\caption{Comparison of relative errors for the Helmholtz equation across different network structures and models}
\resizebox{\textwidth}{!}{
\begin{tabular}{c|c c c c}
\hline
\textbf{Network Structure} & {\textbf{PINN}} & {\textbf{IA-PINN}} & {\textbf{IAW-PINN}} & {\textbf{I-PINN}} \\
\hline
30 units \& 3 hidden layers & \num{6.54e-01} & \cellcolor{blue!10}\num{1.29e-01} & \num{3.02e-01} & \cellcolor{blue!40}\num{1.11e-02} \\
50 units \& 3 hidden layers & \num{1.59e-01} & \num{4.29e-02} & \cellcolor{blue!10}\num{1.32e-02} & \cellcolor{blue!40}\num{7.75e-03} \\
70 units \& 3 hidden layers & \num{1.60e-01} & \num{1.55e-02} & \cellcolor{blue!10}\num{1.14e-02} & \cellcolor{blue!40}\num{5.36e-03} \\
90 units \& 3 hidden layers & \num{9.11e-02} & \num{1.41e-02} & \cellcolor{blue!10}\num{9.19e-03} & \cellcolor{blue!40}\num{4.37e-03} \\
\hline
30 units \& 5 hidden layers & \num{2.11e-01} & \cellcolor{blue!10}\num{4.52e-02} & \num{3.68e-01} & \cellcolor{blue!40}\num{6.74e-03} \\
50 units \& 5 hidden layers & \num{1.01e-01} & \cellcolor{blue!10}\num{1.15e-02} & \num{5.76e-01} & \cellcolor{blue!40}\num{6.60e-03} \\
70 units \& 5 hidden layers & \num{1.50e-01} & \cellcolor{blue!10}\num{1.32e-02} & \num{2.38e-02} & \cellcolor{blue!40}\num{3.49e-03} \\
90 units \& 5 hidden layers & \num{1.35e-01} & \num{9.86e-03} & \cellcolor{blue!10}\num{8.28e-03} & \cellcolor{blue!40}\num{3.77e-03} \\
\hline
30 units \& 7 hidden layers & \num{2.85e-01} & \cellcolor{blue!10}\num{4.55e-02} & \num{1.04e-01} & \cellcolor{blue!40}\num{7.62e-03} \\
50 units \& 7 hidden layers & \num{2.68e-01} & \num{2.22e-02} & \cellcolor{blue!10}\num{1.21e-02} & \cellcolor{blue!40}\num{5.26e-03} \\
70 units \& 7 hidden layers & \num{1.33e-01} & \num{1.80e-02} & \cellcolor{blue!10}\num{8.40e-03} & \cellcolor{blue!40}\num{4.30e-03} \\
90 units \& 7 hidden layers & \num{4.10e-02} & \num{2.29e-02} & \cellcolor{blue!10}\num{1.55e-02} & \cellcolor{blue!40}\num{3.73e-03} \\
\hline
\end{tabular}
}
\label{tab:relative_errors}
\end{table}

\begin{table}[h]
\centering
\caption{Relative errors of different models at various  k  values after 40,000 optimizations using Adam}
\resizebox{0.8\textwidth}{!}{
\begin{tabular}{|c|c|c|c|c|}
\hline
\diagbox[innerwidth=3cm]{\quad k}{Method} &  PINN  & IA-PINN  & IAW-PINN & I-PINN  \\
\hline
$k=2$ & 5.00e-01 & \cellcolor{blue!10}8.43e-02 & 1.72e+00 & \cellcolor{blue!40}3.04e-02 \\
\hline
$k=3$ & 2.39e-01 &\cellcolor{blue!10}1.73e-02 & 1.90e-01 & \cellcolor{blue!40}4.28e-03 \\
\hline
$k=4$ & 7.09e-01 & \cellcolor{blue!10}3.31e-02 & 6.29e-01 & \cellcolor{blue!40}3.80e-03 \\
\hline
\end{tabular}
}
\end{table}

\begin{figure}[htbp]
    \centering
    \includegraphics[width=0.8\linewidth]{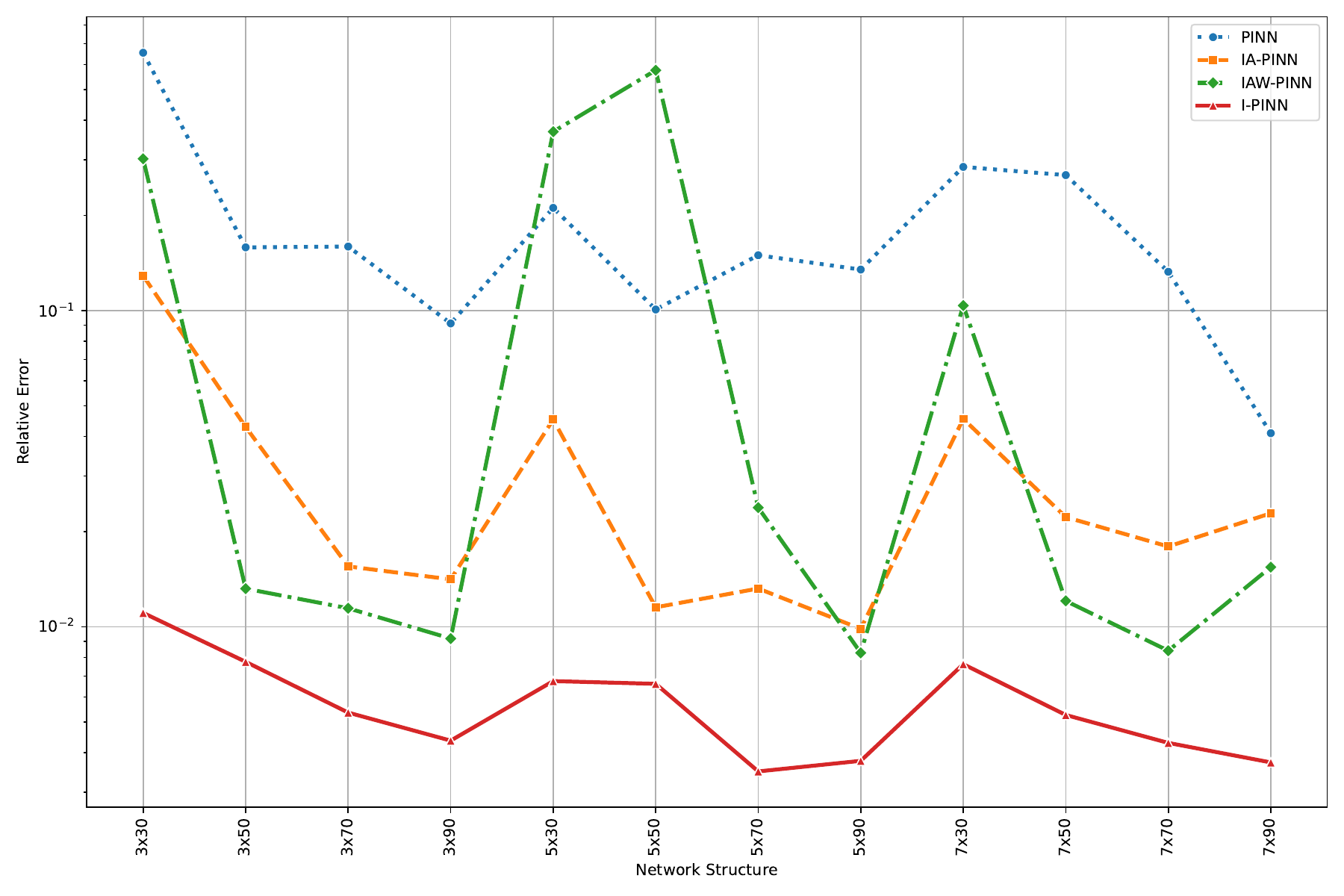}
    \caption{Relative error comparison for the Helmholtz equation across different network structures and models}
    \label{fig:Helmholtz_m}
\end{figure}

Through Figure \ref{fig:Helmholtz_m}, it is evident that I-PINN exhibits superior performance compared to PINN, IA-PINN, and IAW-PINN across various network structures. Typically, its accuracy improves by an order of magnitude compared to IAW-PINN and IA-PINN. Moreover, Figure \ref{fig:Helmholtz_m} indicates that IAW-PINN is less effective in solving problems with relatively simple network structures, sometimes performing even worse than PINN. However, with adjusted network structures, I-PINN consistently demonstrates strong performance and robustness across all scenarios.

\subsection{Klein–Gordon equation}
The Klein-Gordon equation, a fundamental equation in relativistic quantum mechanics and quantum field theory, is a nonlinear partial differential equation of significant importance in scientific disciplines such as quantum mechanics, particle physics, and nonlinear optics \cite{MR2372135}. The equation is formulated as follows:
\begin{equation}\label{kl}
    \left\{\begin{array}{ll}u_{tt}+\alpha\Delta u+\beta u+\gamma u^k=f(x,t),&(x,t)\in\Omega\times[0,T],\\u(x,0)=g_1(x), &x \in\Omega, \\u_t(x,0)=g_2(x), &x \in\Omega,\\u(x,t)=h(x,t), &(x,t)\in\partial\Omega\times[0,T],\end{array}\right.
\end{equation}
where \( \alpha, \beta, \gamma \), and \( k \) are known constants representing Laplace operators acting on spatial variables. The parameter \( k \) denotes the nonlinearity order of the equation. The functions \( f(x, t)\), \(g_1(x)\), \(g_2(x), \) and \( h(x) \) are assumed to be known, while \( u(x, t) \) represents the sought solution. For this study, we define \( \Omega = [0, 1] \times [0, 1] \), \( T = 1 \), \( \alpha = -1 \), \( \beta = 0 \), \( \gamma = 1 \), and \( k = 3 \). The initial conditions are specified with \( g_1(x) = g_2(x) = 0 \) for \( x \in \Omega \). To assess the model's accuracy, we will use a constructed solution:
\begin{equation}
    u(x,t)=x\cos(5\pi t)+(xt)^3
\end{equation}
The external forcing term $f(x,t)$ as well as the initial condition and the Delicacy boundary condition $h(x,y)$ can be obtained from  (\ref{kl}). 

\begin{figure}[htbp]
    \centering
    \subfloat[PINN]{
        \includegraphics[width=\textwidth]{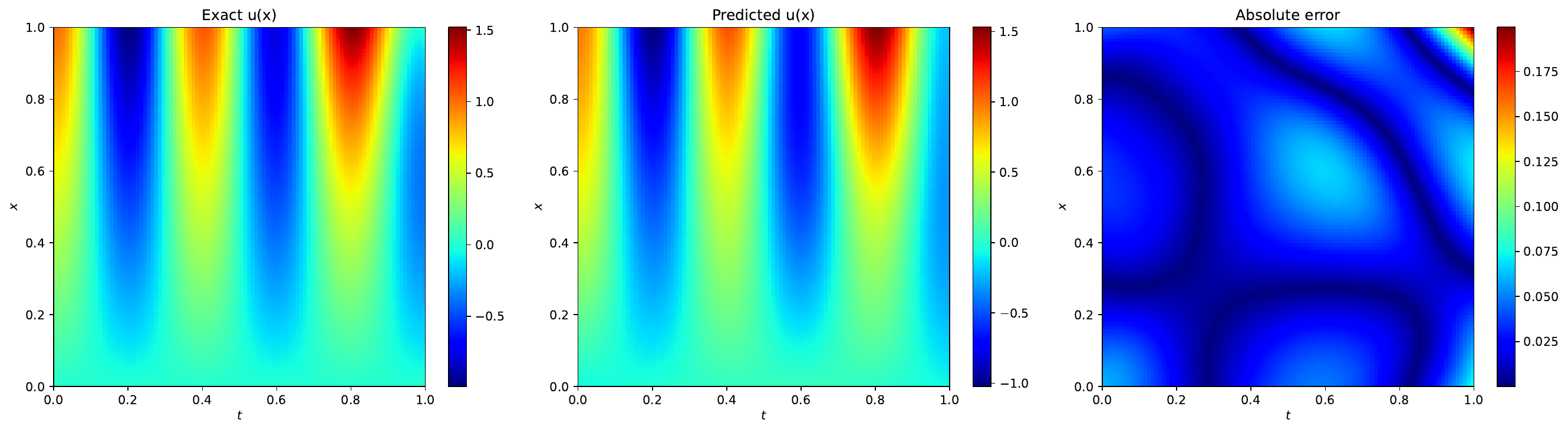}
    }\\
    \subfloat[IA-PINN]{
        \includegraphics[width=\textwidth]{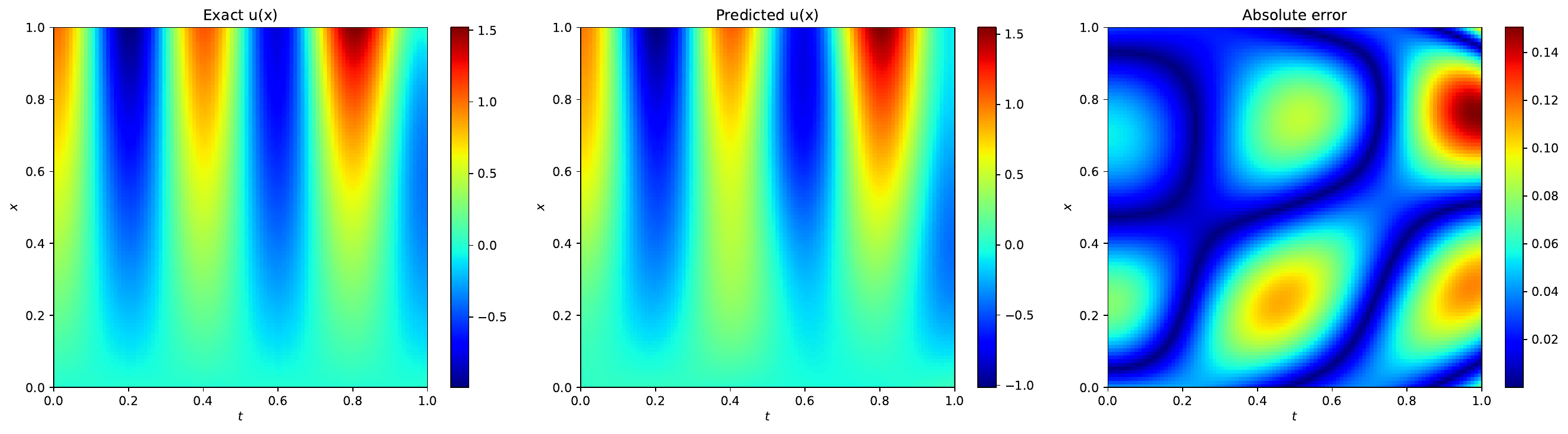}
    }\\
    \subfloat[IAW-PINN]{
        \includegraphics[width=\textwidth]{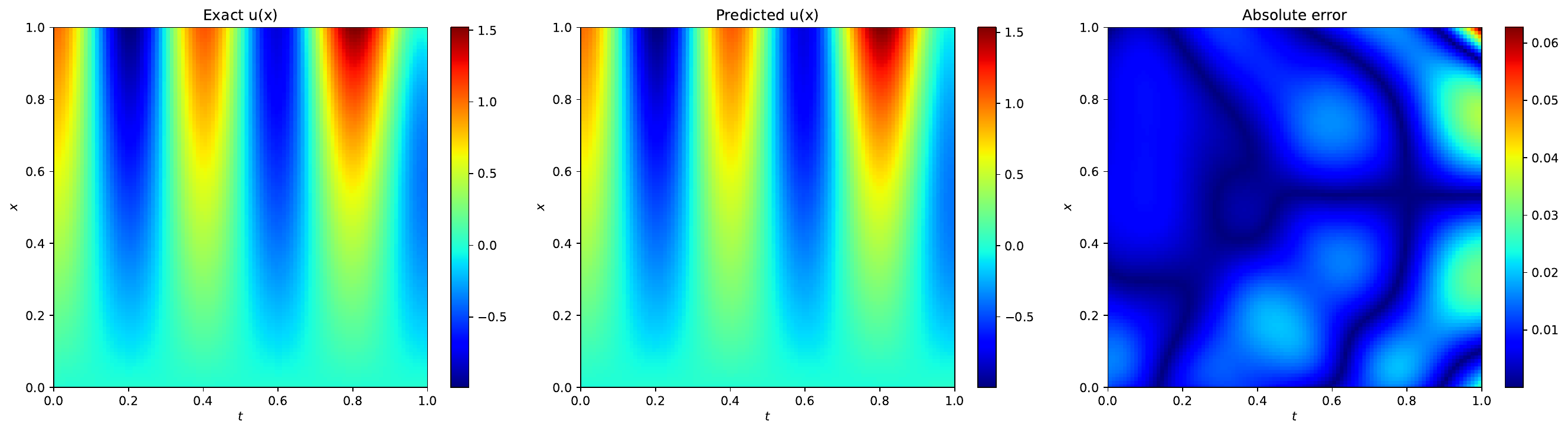}
    }\\
    \subfloat[I-PINN]{
        \includegraphics[width=\textwidth]{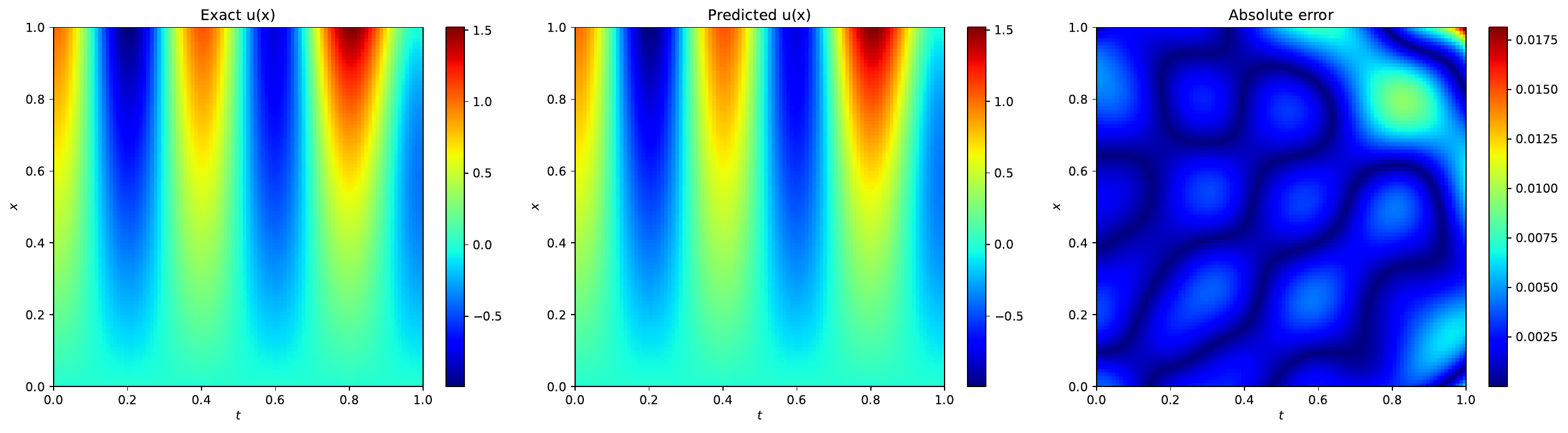}
    }
    \caption{From top to bottom, the figures depict the exact solution, predicted solutions, and absolute errors of the Klein-Gordon equation obtained by employing Adam optimization over 40,000 iterations for the models PINN, IA-PINN, IAW-PINN, and I-PINN, using a network with a 7x50 hidden layer structure.}
    \label{fig:Klein-Gordon—equation}
\end{figure}

In this study, we employed a network architecture with hidden layers of size \( 7 \times 50 \) and set the weight upper bound \(\gamma\) to \( 10^6 \). The Adam optimizer was used to train PINN, IA-PINN, IAW-PINN, and I-PINN models over 40,000 iterations. The configuration of boundary and residual points followed Section 3.1. The resulting relative errors were: 7.21e-2, 1.26e-1, 2.57e-2, 6.70e-3. Clearly, I-PINN demonstrates superior solution accuracy, achieving at least an order of magnitude improvement in relative error compared to the other three methods. Figure  \ref{fig:Klein-Gordon—equation} depicts the exact solution, predicted solution, and absolute error heatmap for I-PINN.

\begin{table}[h]
\centering
\caption{Relative error comparison of the Klein-Gordon equation across various network structures and models}
\resizebox{\textwidth}{!}{
\begin{tabular}{c|c c c c}
\hline
\textbf{Network Structure} & \textbf{PINN} & \textbf{IA-PINN} & \textbf{IAW-PINN} & \textbf{I-PINN} \\
\hline
30 units \& 3 hidden layers & \num{2.65e-1} & \cellcolor{blue!10}\num{4.48e-2} & \num{4.74e-1} & \cellcolor{blue!40}\num{1.43e-2} \\
50 units \& 3 hidden layers & \num{1.92e-1} & \num{7.46e-2} & \cellcolor{blue!10}\num{4.09e-2} & \cellcolor{blue!40}\num{1.01e-2} \\
70 units \& 3 hidden layers & \num{1.44e-1} & \num{5.53e-2} & \cellcolor{blue!10}\num{2.03e-2} & \cellcolor{blue!40}\num{7.50e-3} \\
90 units \& 3 hidden layers & \num{1.06e-1} & \num{5.17e-2} & \cellcolor{blue!10}\num{1.45e-2} & \cellcolor{blue!40}\num{7.30e-3} \\
\hline
30 units \& 5 hidden layers & \num{8.61e-1} & \cellcolor{blue!10}\num{7.33e-2} & \num{5.70e+0} & \cellcolor{blue!40}\num{1.18e-2} \\
50 units \& 5 hidden layers & \num{1.42e-1} & \cellcolor{blue!10}\num{3.95e-2} & \num{5.83e+0} & \cellcolor{blue!40}\num{6.80e-3} \\
70 units \& 5 hidden layers & \num{9.44e-2} & \num{4.37e-2} & \cellcolor{blue!10}\num{2.58e-2} & \cellcolor{blue!40}\num{6.40e-3} \\
90 units \& 5 hidden layers & \num{6.13e-2} & \num{2.75e-2} & \cellcolor{blue!10}\num{2.00e-2} & \cellcolor{blue!40}\num{9.40e-3} \\
\hline
30 units \& 7 hidden layers & \num{1.45e-1} & \cellcolor{blue!10}\num{1.41e-1} & \num{6.20e-1} & \cellcolor{blue!40}\num{1.53e-2} \\
50 units \& 7 hidden layers & \num{7.21e-2} & \num{1.26e-1} & \cellcolor{blue!10}\num{2.57e-2} & \cellcolor{blue!40}\num{6.70e-3} \\
70 units \& 7 hidden layers & \num{2.25e-1} & \num{6.20e-2} & \cellcolor{blue!10}\num{1.57e-2} & \cellcolor{blue!40}\num{7.90e-3} \\
90 units \& 7 hidden layers & \num{1.46e-1} & \num{6.76e-2} & \cellcolor{blue!10}\num{1.26e-2} & \cellcolor{blue!40}\num{9.00e-3} \\
\hline
\end{tabular}
}
\label{tab:Klein–Gordo—relative_errors}
\end{table}

\begin{figure}[htbp]
    \centering
    \includegraphics[width=0.8\textwidth]{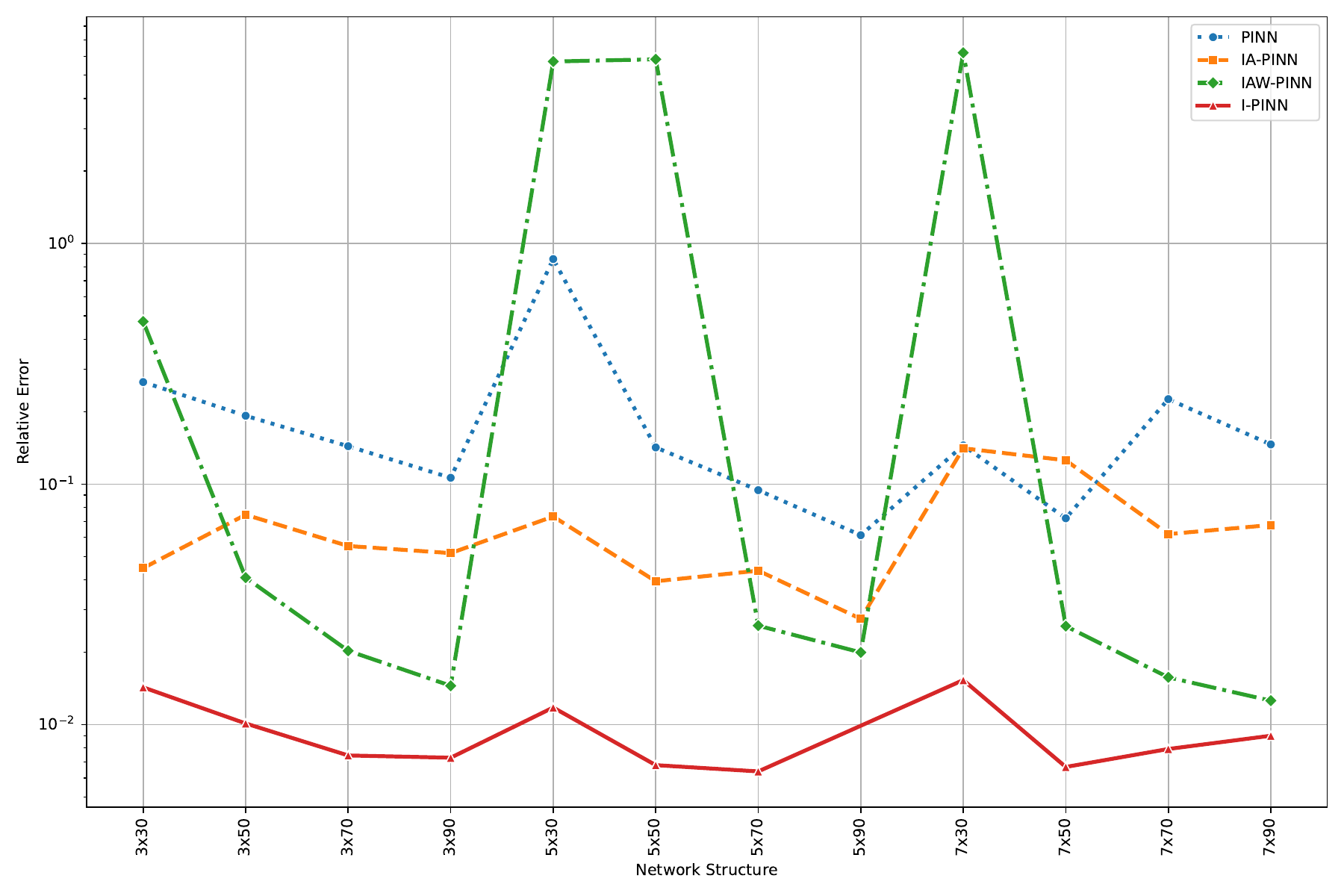}
    \caption{Relative error comparison of the Klein-Gordon Equation across different network structures and models}
    \label{fig:Klein–Gordon_Relative_Error_Comparison_for_Different_Network_Structures_and_Models}
\end{figure}

\begin{figure}[htbp]
    \centering
    \includegraphics[width=0.7\textwidth]{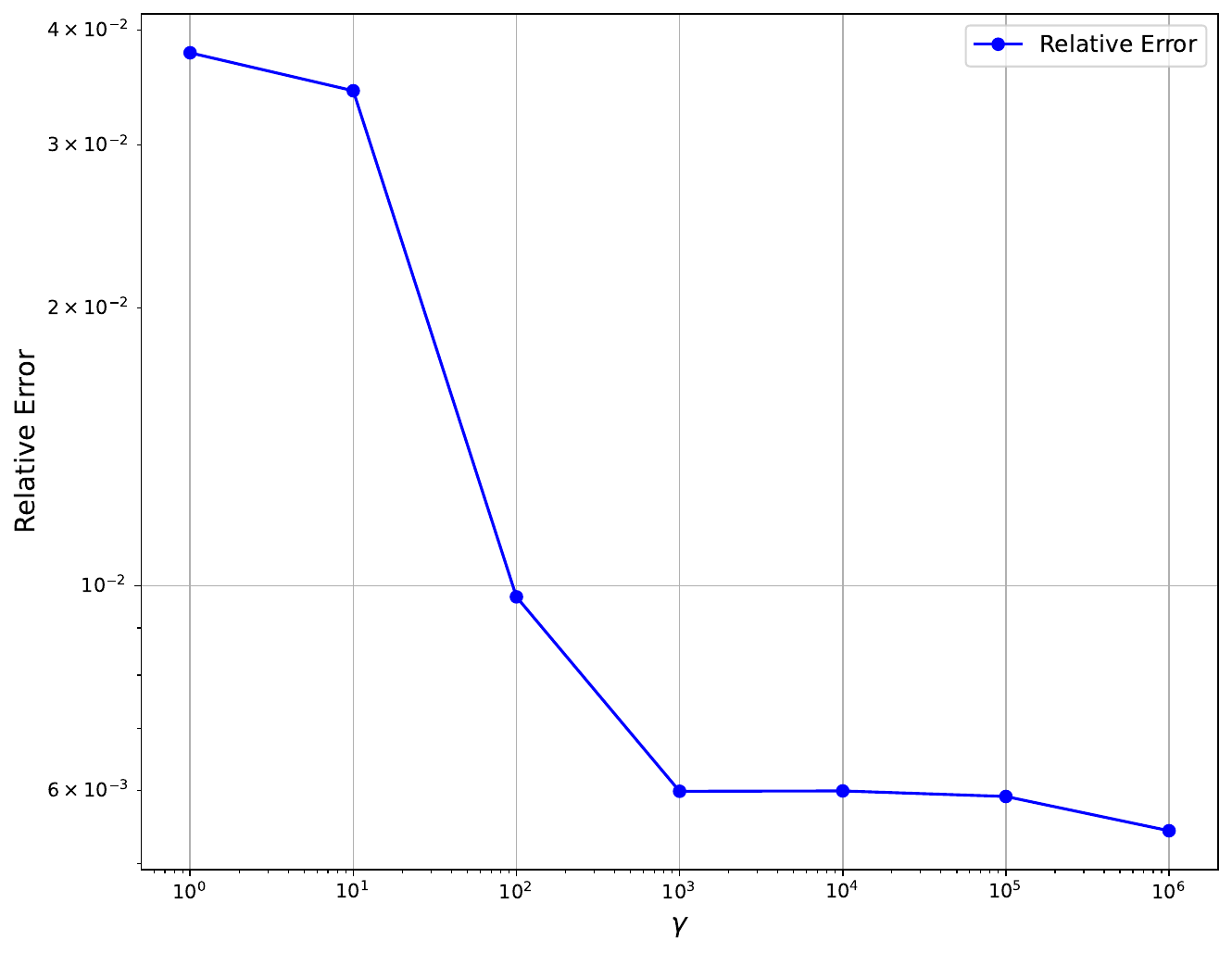}
    \caption{Relative error comparison for I-PINN using a \(7 \times 50\) hidden layer structure with 40,000 iterations of Adam's optimizer under different adaptive weight upper bound cases.}
    \label{fig:Relative——Error——Comparison}
\end{figure}

\begin{figure}[htbp]
    \centering
     \includegraphics[width=0.7\textwidth]{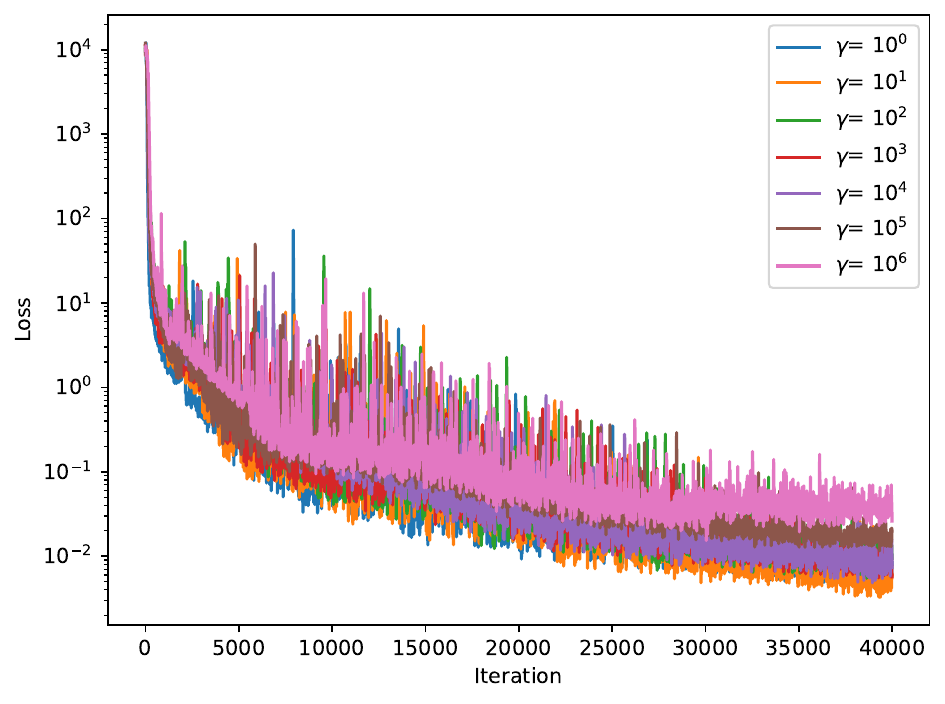}
    \caption{Comparison of loss histories for I-PINN using a \(7 \times 50\) hidden layer structure with 40,000 iterations of Adam's optimizer under different adaptive weight upper bounds.}
    \label{fig:Comparison——of——losses}
\end{figure}

Additionally, we investigated the impact of different adaptive weight caps on the performance of I-PINN under the same network architecture. Various adaptive weight caps were set to analyze their effects on the solution results of I-PINN. Specifically, we employed a network architecture with 7 hidden layers, each containing 50 neurons, and trained the model using the Adam optimizer for 40,000 iterations. The weight upper bounds \(\gamma\) were set to \(10^n\) for (\(n = 0, 1, 2, 3, 4, 5, 6\)), with other settings consistent with Section 3.1. Figure s \ref{fig:Relative——Error——Comparison} and \ref{fig:Comparison——of——losses} present the comparisons of losses and variations in relative errors of I-PINN under different weight caps, respectively. From Figure  \ref{fig:Relative——Error——Comparison}, it is evident that as the weight cap increases, the relative error gradually decreases and tends to level off.

\subsection{Flow in a lid-driven cavity}
Finally, we apply the proposed method to study the problem of lid-driven cavity flow, which is a benchmark problem in computational fluid dynamics with a wide range of applications in engineering and science. In lid-driven flow, the upper boundary of the cavity is assigned a fixed velocity in the $x$ direction, while the other edges of the cavity are stationary and do not move. The incompressible Navier-Stokes equations can be used to analyse this example and can be expressed as follows:
\begin{equation}
    \begin{cases}\boldsymbol{u}\cdot\nabla\boldsymbol{u}+\nabla p=\frac{1}{Re}\Delta\boldsymbol{u}, &\mathrm{in} \Omega\\\nabla\cdot\boldsymbol{u}=0,&\mathrm{in}\Omega\\\boldsymbol{u}(\boldsymbol{x})=(1,0),&\mathrm{on}\Gamma_{1}\\\boldsymbol{u}(\boldsymbol{x})=(0,0),&\mathrm{on}\Gamma_{2}\end{cases}
\end{equation}
In the context of the lid-driven cavity flow problem, where \( u(x) = (u(x), v(x)) \) represents the velocity vector field, \( p \) denotes the scalar pressure field, and \( x = (x, y) \in \Omega = (0, 1) \times (0, 1) \) defines the two-dimensional square cavity \(\Omega\). Here, \(\Gamma_1\) denotes the top boundary of the cavity, \(\Gamma_0\) denotes the other three sides, and \( Re \) is the Reynolds number of the flow. For this illustration, we consider a relatively simple case with \( Re = 100 \). Solving these equations yields insights into the fluid motion within the cavity, including the distribution of velocity and pressure fields.

\begin{figure}[htbp]
    \centering
    \subfloat[PINN]{
        \includegraphics[width=\textwidth]{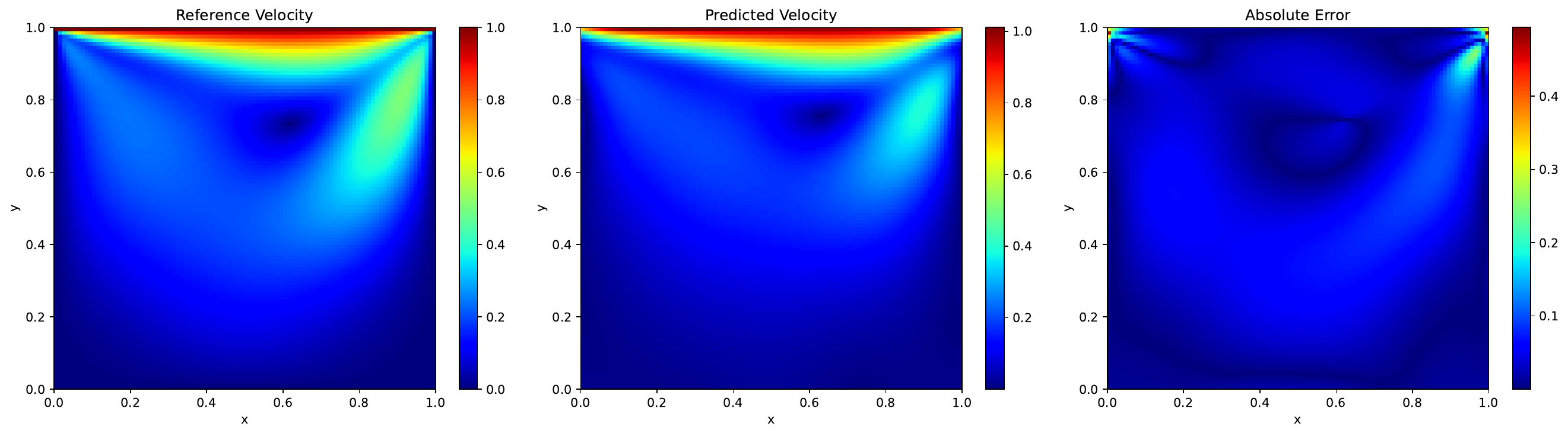}
    }\\
    \subfloat[IA-PINN]{
        \includegraphics[width=\textwidth]{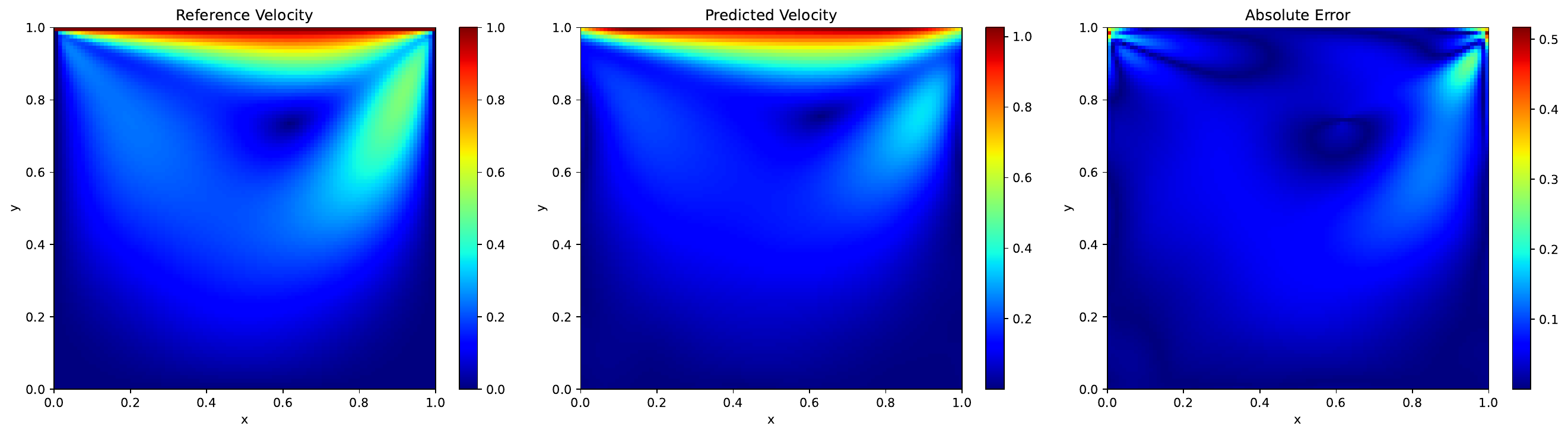}
    }\\
    \subfloat[IAW-PINN]{
        \includegraphics[width=\textwidth]{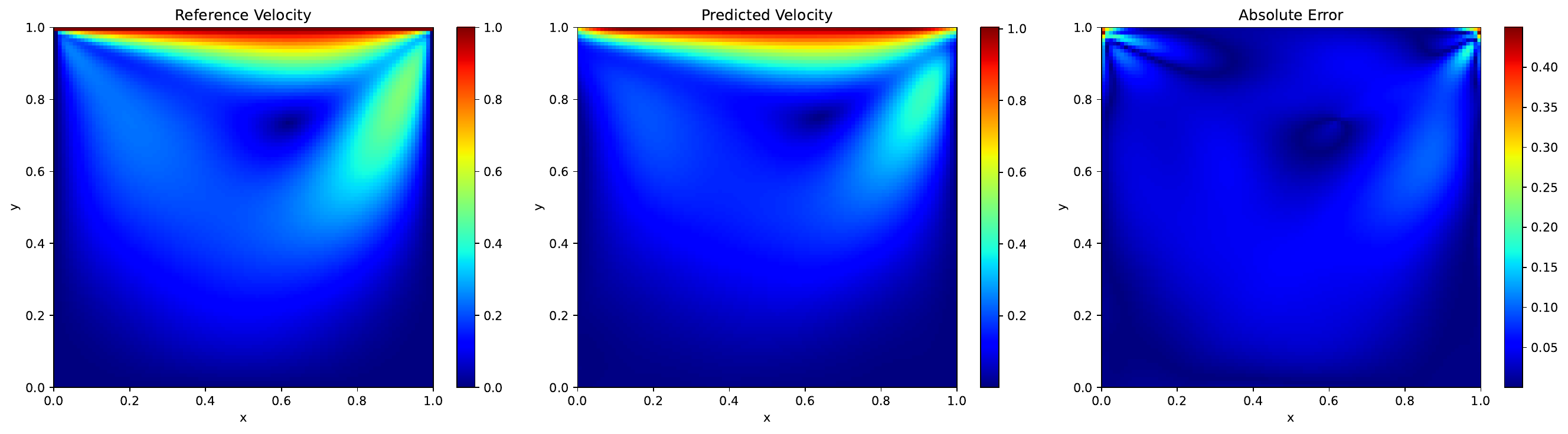}
    }\\
    \subfloat[I-PINN]{
        \includegraphics[width=\textwidth]{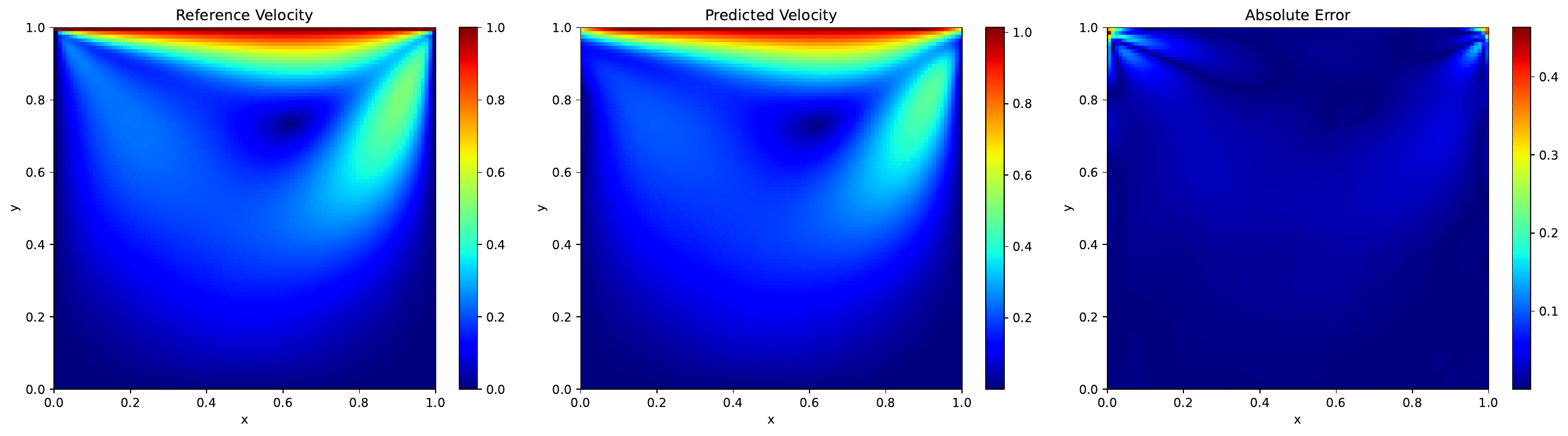}
    }
    \caption{From top to bottom are the reference solution, the predicted solution, and the absolute error of the lid-driven cavity using the PINN, IA-PINN, IAW-PINN, and I-PINN models, respectively. The relative errors for these models are 1.47e-1, 1.07e-1, 1.19e-1, and 5.12e-2.}
    \label{fig:lid-driven}
\end{figure}

In this section, our objective is to learn the velocity components \(u(x,y)\) and \(v(x,y)\), as well as the potential pressure field \(p(x,y)\). We selected a network with 5 hidden layers, each containing 30 neurons, and trained it using the Adam optimizer for 40,000 iterations. The configuration of boundary and residual points was consistent with Section 3.1. Figure  \ref{fig:lid-driven} summarizes the experimental results. Specifically, we compared the reference solutions and predicted solutions obtained from different models. It was observed that the velocity fields obtained by PINN, IA-PINN, and IAW-PINN were very close to the reference solution, with relative errors of 1.47e-1, 1.07e-1, and 1.19e-1, respectively. However, I-PINN achieved highly satisfactory results, with a relative error of 5.12e-02.

\section{Conclusions} \label{sub4}
This research presents the development and evaluation of the I-PINN, a novel approach designed to enhance both the predictive accuracy and the convergence of solutions for PDEs. The I-PINN model has demonstrated superior performance, exceeding the accuracy of PINN, IAW-PINN, and IA-PINN by at least an order of magnitude, all while maintaining computational efficiency.

The key findings of this study are as follows: I-PINN exhibits robust performance across various network structures and problem complexities, demonstrating its strong generalization capabilities. Additionally, I-PINN's innovative framework integrates an enhanced architecture with improved adaptive weighting, resulting in stable and convergent optimization for PDEs.
 
While I-PINN signifies a significant advancement in the field of PINNs, providing a robust and accurate solution strategy for PDEs, further research is necessary to explore its potential applications across diverse disciplines.

\begin{enumerate}
\item The selection of upper bounds for the weights in IAW-PINN remains an open question; it is an empirical choice, and the selection of the weight upper bound \(\gamma\) may vary for different problems. To address this, we will examine the combination of min-max optimization and an improved architecture as an alternative to weight upper bounds in PINNs.
\item Further exploration of adaptive weighting strategies to balance loss components in PINNs is necessary. Additionally, we will investigate the scalability of I-PINN to higher-dimensional PDEs and more complex physical systems.
\item We aim to extend the I-PINN framework to handle multi-scale and multi-physics problems, enhancing its applicability in real-world engineering and scientific challenges.
\item The utilization of I-PINN in the context of inverse problems also requires exploration, aiming to broaden its functionality across various problem-solving domains.
\end{enumerate}

\textbf{Funding information}: The authors extend their appreciation to the supported by  the National Natural Science Foundation of China No.11971337; the Scientific
Research Foundation of Chengdu University of Information Technology No.KYQN202324; the Key Laboratory
of Numerical Simulation of Sichuan Provincial Universities No.KLNS-2023SZFZ002;
the Science and Technology Department of Sichuan Province No.2022JDR0043.

\textbf{Conflict of interest}: The authors declare no potential conflict of interest.

\bibliographystyle{siam}
\bibliography{myfile}

\begin{thebibliography}{10}

\bibitem{MR4698526}
{\sc S.~J. Anagnostopoulos, J.~D. Toscano, N.~Stergiopulos, and G.~E.
  Karniadakis}, {\em Residual-based attention in physics-informed neural
  networks}, Comput. Methods Appl. Mech. Engrg., 421 (2024), pp.~Paper No.
  116805, 19.

\bibitem{MR4425212}
{\sc S.~Basir and I.~Senocak}, {\em Physics and equality constrained artificial
  neural networks: application to forward and inverse problems with
  multi-fidelity data fusion}, J. Comput. Phys., 463 (2022), pp.~Paper No.
  111301, 18.

\bibitem{Cipolla_Gal_Kendall_2018}
{\sc R.~Cipolla, Y.~Gal, and A.~Kendall}, {\em Multi-task learning using
  uncertainty to weigh losses for scene geometry and semantics}, in 2018
  IEEE/CVF Conference on Computer Vision and Pattern Recognition, Jun 2018.

\bibitem{MR4764284}
{\sc T.~G. Grossmann, U.~J. Komorowska, J.~Latz, and C.-B. Sch\"{o}nlieb}, {\em
  Can physics-informed neural networks beat the finite element method?}, IMA J.
  Appl. Math., 89 (2024), pp.~143--174.

\bibitem{MR4426047}
{\sc J.~M. Hanna, J.~V. Aguado, S.~Comas-Cardona, R.~Askri, and
  D.~Borzacchiello}, {\em Residual-based adaptivity for two-phase flow
  simulation in porous media using physics-informed neural networks}, Comput.
  Methods Appl. Mech. Engrg., 396 (2022), pp.~Paper No. 115100, 16.

\bibitem{Hou2023}
{\sc J.~Hou, Y.~Li, and S.~Ying}, {\em Enhancing pinns for solving pdes via
  adaptive collocation point movement and adaptive loss weighting}, Nonlinear
  Dyn., 111 (2023), pp.~15233--15261.

\bibitem{Hu_2023}
{\sc Z.~Hu, A.~D. Jagtap, G.~E. Karniadakis, and K.~Kawaguchi}, {\em Augmented
  physics-informed neural networks (apinns): A gating network-based soft domain
  decomposition methodology}, Engineering Applications of Artificial
  Intelligence, 126 (2023), p.~107183.

\bibitem{arXiv211101394}
{\sc X.~Huang, H.~Liu, B.~Shi, Z.~Wang, K.~Yang, Y.~Li, B.~Weng, M.~Wang,
  H.~Chu, J.~Zhou, F.~Yu, B.~Hua, L.~Chen, and B.~Dong}, {\em Solving partial
  differential equations with point source based on physics-informed neural
  networks}, arXiv:2111.01394,  (2021).

\bibitem{MR4188528}
{\sc A.~D. Jagtap and G.~E. Karniadakis}, {\em Extended physics-informed neural
  networks ({XPINN}s): a generalized space-time domain decomposition based deep
  learning framework for nonlinear partial differential equations}, Commun.
  Comput. Phys., 28 (2020), pp.~2002--2041.

\bibitem{MR4051868}
{\sc A.~D. Jagtap, K.~Kawaguchi, and G.~E. Karniadakis}, {\em Adaptive
  activation functions accelerate convergence in deep and physics-informed
  neural networks}, J. Comput. Phys., 404 (2020), pp.~109136, 23.

\bibitem{MR4133779}
\leavevmode\vrule height 2pt depth -1.6pt width 23pt, {\em Locally adaptive
  activation functions with slope recovery for deep and physics-informed neural
  networks}, Proc. A., 476 (2020), pp.~20200334, 20.

\bibitem{MR4083367}
{\sc A.~D. Jagtap, E.~Kharazmi, and G.~E. Karniadakis}, {\em Conservative
  physics-informed neural networks on discrete domains for conservation laws:
  applications to forward and inverse problems}, Comput. Methods Appl. Mech.
  Engrg., 365 (2020), pp.~113028, 27.

\bibitem{Karniadakis2021}
{\sc G.~E. Karniadakis, I.~G. Kevrekidis, L.~Lu, P.~Perdikaris, S.~Wang, and
  L.~Yang}, {\em Physics-informed machine learning}, Nat. Rev. Phys., 3 (2021),
  pp.~422--440.

\bibitem{arXiv170507115}
{\sc A.~Kendall, Y.~Gal, and R.~Cipolla}, {\em Multi-task learning using
  uncertainty to weigh losses for scene geometry and semantics},
  arXiv:1705.07115,  (2018).

\bibitem{MR4550083}
{\sc L.~R. Lima and L.~L. Godeiro}, {\em Equity-premium prediction: attention
  is all you need}, J. Appl. Econometrics, 38 (2023), pp.~105--122.

\bibitem{LIU2021112}
{\sc D.~Liu and Y.~Wang}, {\em A dual-dimer method for training
  physics-constrained neural networks with minimax architecture}, Neural
  Networks, 136 (2021), pp.~112--125.

\bibitem{MR4209661}
{\sc L.~Lu, X.~Meng, Z.~Mao, and G.~E. Karniadakis}, {\em Deep{XDE}: a deep
  learning library for solving differential equations}, SIAM Rev., 63 (2021),
  pp.~208--228.

\bibitem{arXiv210204626}
{\sc L.~Lu, R.~Pestourie, W.~Yao, Z.~Wang, F.~Verdugo, and S.~G. Johnson}, {\em
  Physics-informed neural networks with hard constraints for inverse design},
  2021.

\bibitem{MR4513793}
{\sc L.~D. McClenny and U.~M. Braga-Neto}, {\em Self-adaptive physics-informed
  neural networks}, J. Comput. Phys., 474 (2023), pp.~Paper No. 111722, 23.

\bibitem{MR3881695}
{\sc M.~Raissi, P.~Perdikaris, and G.~E. Karniadakis}, {\em Physics-informed
  neural networks: a deep learning framework for solving forward and inverse
  problems involving nonlinear partial differential equations}, J. Comput.
  Phys., 378 (2019), pp.~686--707.

\bibitem{MR4350501}
{\sc R.~van~der Meer, C.~W. Oosterlee, and A.~Borovykh}, {\em Optimally
  weighted loss functions for solving {PDE}s with neural networks}, J. Comput.
  Appl. Math., 405 (2022), pp.~Paper No. 113887, 18.

\bibitem{MR4309866}
{\sc S.~Wang, Y.~Teng, and P.~Perdikaris}, {\em Understanding and mitigating
  gradient flow pathologies in physics-informed neural networks}, SIAM J. Sci.
  Comput., 43 (2021), pp.~A3055--A3081.

\bibitem{MR4337814}
{\sc S.~Wang, X.~Yu, and P.~Perdikaris}, {\em When and why {PINN}s fail to
  train: a neural tangent kernel perspective}, J. Comput. Phys., 449 (2022),
  pp.~Paper No. 110768, 28.

\bibitem{MR4748739}
{\sc Y.~Wang, Y.~Yao, J.~Guo, and Z.~Gao}, {\em A practical {PINN} framework
  for multi-scale problems with multi-magnitude loss terms}, J. Comput. Phys.,
  510 (2024), p.~Paper No. 113112.

\bibitem{MR2372135}
{\sc A.-M. Wazwaz}, {\em New travelling wave solutions to the {B}oussinesq and
  the {K}lein-{G}ordon equations}, Commun. Nonlinear Sci. Numer. Simul., 13
  (2008), pp.~889--901.

\bibitem{MR4499321}
{\sc C.~Wu, M.~Zhu, Q.~Tan, Y.~Kartha, and L.~Lu}, {\em A comprehensive study
  of non-adaptive and residual-based adaptive sampling for physics-informed
  neural networks}, Comput. Methods Appl. Mech. Engrg., 403 (2023), pp.~Paper
  No. 115671, 23.

\bibitem{Xiang_Peng_Liu_Yao_2022}
{\sc Z.~Xiang, W.~Peng, X.~Liu, and W.~Yao}, {\em Self-adaptive loss balanced
  physics-informed neural networks}, Neurocomputing,  (2022), p.~11–34.

\bibitem{MR4746527}
{\sc R.~Zhang, G.~P. Warn, and A.~Radli\'{n}ska}, {\em Physics-{I}nformed
  {P}arallel {N}eural {N}etworks with self-adaptive loss weighting for the
  identification of continuous structural systems}, Comput. Methods Appl. Mech.
  Engrg., 427 (2024), p.~Paper No. 117042.

\end{thebibliography}
\end{document}